\documentclass{article} 
\usepackage{iclr2027_conference,times}

\usepackage{amsmath,amsfonts,bm}

\def\eqref#1{equation~\ref{#1}}

\def\1{\bm{1}}

\DeclareMathAlphabet{\mathsfit}{\encodingdefault}{\sfdefault}{m}{sl}
\SetMathAlphabet{\mathsfit}{bold}{\encodingdefault}{\sfdefault}{bx}{n}

\usepackage{hyperref}
\usepackage{url}
\usepackage{graphicx}
\usepackage{multirow}
\usepackage{booktabs}
\usepackage{colortbl}
\usepackage{algorithm}
\usepackage{algpseudocode}
\usepackage[most]{tcolorbox}

\newtcolorbox{promptbox}[1]{
  breakable,
  colback=gray!3,
  colframe=gray!55,
  boxrule=0.5pt,
  arc=1mm,
  left=6pt,
  right=6pt,
  top=5pt,
  bottom=5pt,
  fonttitle=\bfseries,
  fontupper=\small,
  title={#1}
}

\title{See, Measure, and Reason: Learning Visually Grounded Reasoning in Pathology}

\author{%
\textbf{Chengyang Zhang\textsuperscript{1,2},
Wenchuang Zhang\textsuperscript{2},
Bo Li\textsuperscript{3},
Mengran Li\textsuperscript{4}
}\\[-1pt]
\textbf{Xinyu Liu\textsuperscript{1},
Jiaming Yang\textsuperscript{1},
Jie Chen\textsuperscript{2},
Zhang Zhang\textsuperscript{2}
}\\[-1pt]
\textbf{Yuhao Yi\textsuperscript{1,2}\thanks{Corresponding author},
Hong Bu\textsuperscript{2},
Jiancheng Lv\textsuperscript{1}
}\\[2pt]
\normalfont
\textsuperscript{1}College of Computer Science, Sichuan University\\[-1pt]
\textsuperscript{2}Department of Pathology and Institute of Clinical Pathology, West China Hospital, Sichuan University\\[-1pt]
\textsuperscript{3}Department of Computer Science, School of Computing, National University of Singapore\\[-1pt]
\textsuperscript{4}School of Intelligent Systems Engineering, Sun Yat-sen University\\
\texttt{yuhaoyi@scu.edu.cn} \\
}

\iclrfinalcopy

\begin{document}

\maketitle
\fancyhead{}
\thispagestyle{fancy}
\begin{abstract}
Pathological assessment relies on recognizing fine-grained visual details in histological images. Vision-language models (VLMs) increasingly support pathology interpretation, yet their ability to perceive these details remains inadequate. This weakness leads to inaccurate cellular observations that can persist even when final answers are correct. In this paper, we propose ASPECT to improve visually grounded reasoning through explicit supervision of cellular appearance and abundance. ASPECT trains intermediate visual tokens through pathology feature reconstruction, cell feature alignment, and count supervision. Three-stage supervised fine-tuning teaches the model to perceive, generate visual tokens, and reason, followed by reinforcement learning that rewards answer correctness and consistency with reported measurements. We also introduce PathoVernier, a benchmark of 759 expert-reviewed questions from five pathology datasets covering four cellular composition tasks. It evaluates both final answers and intermediate measurements to expose errors hidden by answer accuracy. On PathoVernier, ASPECT achieves relative accuracy gains of approximately 19.2\% over the strongest baseline, Gemini-3.1-Pro, and 99.3\% over its Qwen3-VL-8B backbone, while reducing RAWR, which measures counting errors within correct responses, by 28.1\% and 42.7\%, respectively. ASPECT also improves over its backbone on three external pathology benchmarks covering classification and question answering beyond cellular composition tasks.
\end{abstract}

\section{Introduction}
\label{sec:intro}
Pathology vision-language models (VLMs) can describe tissue morphology and answer diagnostic questions from histological images \citep{lu2024multimodal,seyfioglu2024quilt,sun2025pathgen,zhang2026patho}. However, success in question answering does not establish how accurately a model identifies the visual details needed for its answers. Recent pathology evaluations reveal difficulties with localization, counting, and spatial relationships, even on slides where models answer higher-level questions correctly \citep{chen2026pathview,zhang2026pathology}. These findings motivate a closer examination of the visual understanding underlying pathology VLM predictions.

General-purpose VLM evaluations examine fine-grained perception through spatial judgments and counting \citep{fu2024blink,rahmanzadehgervi2024vision,zhou2025they}. In pathology, cellular composition analysis requires these abilities and supports subsequent tissue assessment. Tumor content informs specimen suitability for molecular testing \citep{lindeman2018updated}, while assessment of tumor-infiltrating lymphocytes requires identifying inflammatory cells within the appropriate tissue compartment \citep{salgado2015evaluation}. Distinguishing cell types and comparing their abundance across densely populated regions therefore provide a clinically relevant test of visual understanding. Nucleus instance annotations supply reference identities and positions, allowing both the reported measurements and the conclusions drawn from them to be checked.

\begin{figure*}[t!]
  \setlength{\abovecaptionskip}{0pt}
  \includegraphics[width=1.\textwidth]{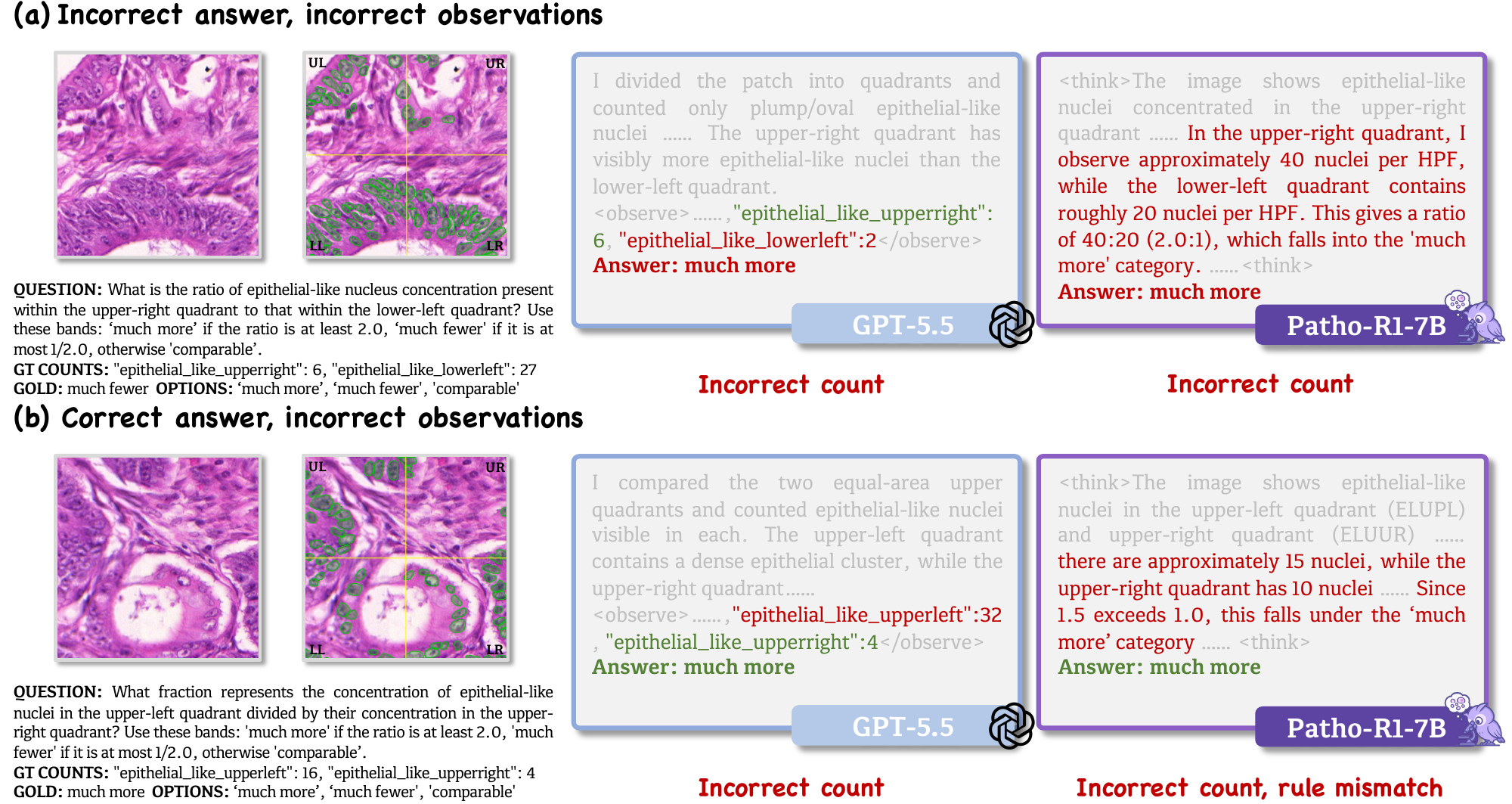}
  \caption{Correct answers can conceal inaccurate observations. GPT-5.5 and Patho-R1 reverse regional nuclear densities in (a) and answer correctly despite incorrect counts in (b), where Patho-R1 also misapplies the ratio rule. Green contours mark reference nucleus annotations.}
  \label{fig1}
\end{figure*}

Our evaluation reveals errors in both observing cellular composition and using those observations to answer questions. Figure \ref{fig1} illustrates these failures in GPT-5.5 \citep{openai2026gpt55} and Patho-R1 \citep{zhang2026patho}, representing a frontier closed-source VLM and a pathology-tuned model. In Figure \ref{fig1}(a), both models reverse the relative epithelial-like nuclear densities of two regions. In Figure \ref{fig1}(b), both select the correct answer despite inaccurate counts; Patho-R1 additionally selects a category inconsistent with the ratio implied by its own counts. \textit{Thus, a correct answer can hide both inaccurate observation and incorrect use of the stated evidence.} We hypothesize that fine-tuning focused on answer generation can strengthen pathology-specific language priors without comparable gains in extracting and using visual information. Consistent with this hypothesis, pathology-tuned models retain substantial accuracy without images, and their gains over base models need not coincide with better visual grounding \citep{zhang2026pathology}. Similar image-independent performance has been reported on broader medical multimodal benchmarks \citep{asadi2026mirage}. These findings motivate evaluating the observations behind an answer and training models to make accurate observations and use them correctly.

To connect fine-grained perception with pathology reasoning, we introduce ASPECT and PathoVernier. ASPECT supervises intermediate visual tokens with pathology features and cell counts, using three-stage supervised fine-tuning (SFT) to generate quantitative observations and reason from them. Reinforcement learning (RL) further rewards answer correctness and consistency with the reported counts. PathoVernier provides 759 expert-reviewed questions from five datasets spanning more than 15 organs. Four cellular composition tasks pair each answer with reference counts and a decision rule, allowing evaluation beyond final-answer accuracy. Our main contributions are summarized as follows:
\begin{itemize}
    \item We introduce PathoVernier to assess fine-grained pathology perception through four cellular composition tasks. Its RAWR metric (right answer, wrong reason) exposes counting errors hidden by correct answers.
    \item We propose ASPECT, which trains intermediate visual tokens to encode pathology appearance and cell abundance through three-stage SFT, followed by RL for answer correctness and consistency with reported counts.
    \item We conduct experiments against ten baselines. ASPECT-8B leads on all PathoVernier metrics, improving accuracy by 19.2\% and reducing RAWR by 28.1\% relative to the strongest baseline. Gains on three external benchmarks extend beyond cellular composition tasks.
\end{itemize}

\section{Related Work}
\label{sec:relworks}
\paragraph{Pathology Vision-Language Models} Pathology VLMs combine domain-specific visual representations with language supervision to describe histological findings and answer diagnostic questions \citep{lu2024multimodal,seyfioglu2024quilt,sun2025pathgen}. PathChat combines a pathology vision encoder with language instruction tuning~\citep{lu2024multimodal}. Reasoning-oriented models subsequently introduce chain-of-thought supervision and reinforcement learning \citep{zhang2026patho}. More recently, PathReasoner-R1 uses knowledge-guided reasoning data and entity-based rewards to strengthen the connection between pathological findings and diagnoses \citep{jiang2026pathreasoner}. PathFound seeks additional evidence to refine diagnoses~\citep{hua2026pathfound}. RECAP-PATH optimizes prompts for morphological descriptions using diagnostic feedback~\citep{hong2026adaptive}. However, diagnostic accuracy and plausible explanations do not establish whether reported cell counts are correct or support a quantitative conclusion.

\paragraph{Evaluating Visual Understanding} Visual-understanding benchmarks assess fine-grained perception, visual reasoning, and domain-specific knowledge \citep{fu2024blink,sun2024pathmmu,chen2024we}. General-domain evaluations isolate perceptual operations, whereas pathology benchmarks emphasize morphological recognition and diagnostic reasoning. For example, BLINK evaluates abilities such as visual correspondence, relative depth estimation, and multi-view reasoning \citep{fu2024blink}, while PathMMU assesses expert-level pathology understanding through question answering \citep{sun2024pathmmu}. More targeted pathology evaluations examine image dependence, entity–region correspondence, and visual understanding across local regions and whole-slide views \citep{chen2026pathview,zhang2026pathology}. However, these pathology evaluations largely assess visual abilities through separate task outputs, without jointly checking a final answer and its observations. 

\paragraph{Visual Supervision and Process Verification} Visual supervision and process verification provide complementary learning signals beyond final-answer correctness \citep{qin2025chain,xiao2026perception,li2025self,pronesti2026beyond}. Existing approaches supervise intermediate visual representations or assess a model's observations and reasoning steps. For example, CoVT trains continuous visual tokens to reconstruct features from vision experts \citep{qin2025chain}. More recently, PEARL constructs verifiable perception questions for each reasoning task and uses auxiliary perception rollouts to guide subsequent reasoning updates \citep{zhang2026perceptual}. However, feature reconstruction does not verify the numerical observations stated in a response, and PEARL checks perception through separately generated answers. These objectives do not directly check whether the reported measurements are accurate and support the conclusion within the same reasoning response.

\section{Methods}
\label{sec:method}
\subsection{Overview}
\label{sec:overview}
\begin{figure*}[t!]
  \includegraphics[width=1.\textwidth]{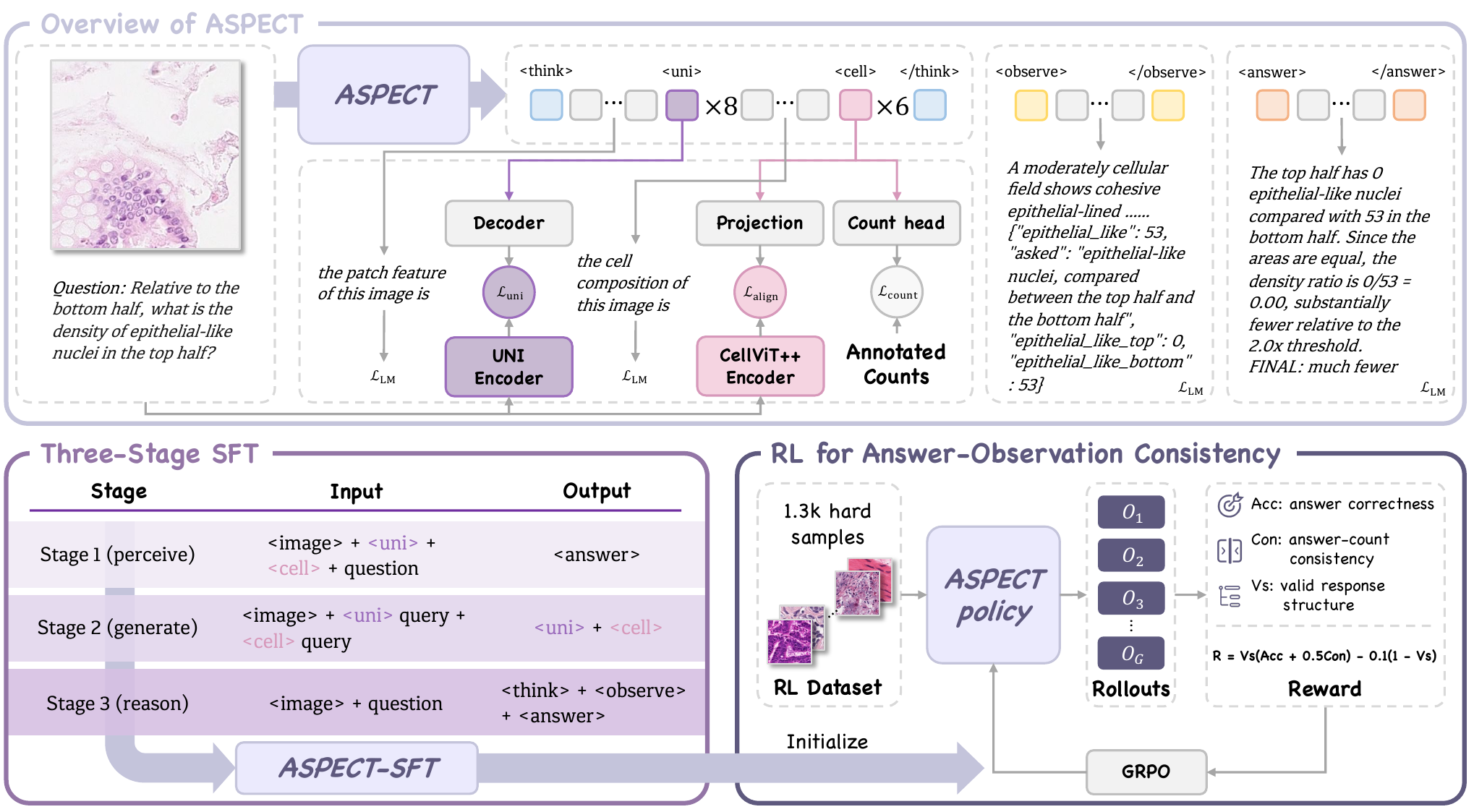}
  \caption{Overview of ASPECT. Pathology feature reconstruction, cell feature alignment, and count supervision train the visual-token representations. Three-stage SFT teaches the model to perceive, generate, and reason with explicit measurements. Subsequent RL rewards answer correctness and consistency with the reported counts.}
  \label{fig2}
\end{figure*}

ASPECT learns to perceive pathology images, report quantitative observations, and reason from them to an answer (Figure \ref{fig2}). Built on Qwen3-VL-8B \citep{bai2025qwen3}, it generates eight pathology feature tokens and six cell tokens before producing the textual observations and answer. We supervise their hidden states through pathology feature reconstruction, cell feature alignment, and count prediction. Three-stage SFT connects these representations to observation and answer generation; RL then rewards correct answers supported by the reported measurements. PathoVernier evaluates this connection through 759 questions spanning four quantitative tasks and more than 15 organs, with reference measurements for every answer.

\subsection{Pathology Visual Supervision}
\label{sec:visual_supervision}
Cell composition analysis requires both recognizing cell appearance and measuring abundance. Supervising intermediate visual tokens can retain image information within autoregressive generation \citep{qin2025chain,li2026latent,yang2026machine}. For composition analysis, these representations must also encode cell identity and abundance. We therefore pair pathology feature reconstruction with cell feature alignment and count supervision, connecting intermediate visual states to the quantitative observations needed for answering. In the complete response, the two token groups, abbreviated as \texttt{<uni>} and \texttt{<cell>}, appear in the \texttt{<think>} block, followed by observations in \texttt{<observe>}, and reasoning with a final answer in \texttt{<answer>}. Let $\bm H_{\mathrm u}\in\mathbb R^{8\times d}$ and $\bm H_{\mathrm c}\in\mathbb R^{6\times d}$ denote the final-layer hidden states at the two groups, where $d$ is the language model's hidden dimension. The following losses train these states to encode the visual information used in subsequent observation and answer generation.

\paragraph{Pathology feature reconstruction.}
To preserve pathology appearance in the \texttt{<uni>} states, we reconstruct features $\bm F_{\mathrm{UNI}}(x)$ extracted from image $x$ by a frozen UNI encoder \citep{chen2024towards}. A cross-attention decoder takes learned queries $\bm U$ and the projected token states as inputs:
\begin{equation}
\begin{aligned}
\widehat{\bm F}_{\mathrm u}
=\operatorname{Dec}_{\mathrm u}
\bigl(\bm U,\operatorname{norm}(\bm H_{\mathrm u}\bm W_{\mathrm u})\bigr),\quad
\mathcal L_{\mathrm{uni}}
=\left\|\widehat{\bm F}_{\mathrm u}
-\bm F_{\mathrm{UNI}}(x)\right\|_F^2.
\end{aligned}
\label{eq:uni_reconstruction}
\end{equation}
Here, $\widehat{\bm F}_{\mathrm u}$ denotes the reconstructed features, $\bm W_{\mathrm u}$ is a learned projection, $\operatorname{norm}$ applies $L_2$ normalization, and $\|\cdot\|_F$ is the Frobenius norm. The cross-attention decoder uses shared learned queries $\bm U$ and the projected states as keys and values, so image-specific information must pass through $\bm H_{\mathrm u}$.

\paragraph{Cell feature alignment.}
With guidance from pathologists, we map source annotations to five target categories: tumor cells, normal epithelial cells, stromal-like cells, lymphocytes, and other inflammatory cells. Finer subtypes within a target category are merged; coarse labels spanning several targets retain their group membership. The first five \texttt{<cell>} tokens represent these categories, and the sixth summarizes all retained, mappable instances. A frozen CellViT++ model~\citep{horst2026cellvit++} supplies instance features $\bm u_j\in\mathbb R^{d_T}$, where $d_T$ is the teacher feature dimension. We pool features within annotated masks or predicted instances; NuCLS contributes counts but no alignment targets. For token $s\in\{1,\ldots,6\}$, let $\mathcal J_s$ index its contributing instances. For a non-empty group, its target $\bm t_s$ averages individually normalized features:
\begin{equation}
\overline{\bm u}_j
=\frac{\bm u_j}{\|\bm u_j\|_2},
\qquad
\bm t_s
=\frac{1}{|\mathcal J_s|}
\sum_{j\in\mathcal J_s}\overline{\bm u}_j.
\label{eq:cell_teacher}
\end{equation}

An instance with a coarse label contributes to each corresponding category target: an inflammatory instance, for example, supervises both lymphocyte and other-inflammatory features. The sixth token pools each retained instance once; excluded labels are listed in Appendix~\ref{app:category_mapping}. We align token states to these targets through
\begin{equation}
\mathcal L_{\mathrm{align}}=\frac{1}{d_T|\mathcal V|}\sum_{s\in\mathcal V}\left\|\bm W_{\mathrm c}(\bm h_s+\bm r_s)-\bm t_s\right\|_2^2,
\label{eq:cell_alignment}
\end{equation}
where $\bm h_s\in\mathbb R^d$ is the transposed $s$-th row of $\bm H_{\mathrm c}$, $\bm r_s\in\mathbb R^d$ is a learned role embedding, and $\bm W_{\mathrm c}\in\mathbb R^{d_T\times d}$ is a learned projection. The set $\mathcal V=\{s:|\mathcal J_s|>0\}$ excludes empty instance groups.

\paragraph{Count supervision.}

Mean pooling summarizes appearance without explicitly preserving instance counts. We therefore train the cell-token states to predict the image-wide abundance of the five target categories. An MLP $g$ predicts the five image-wide category counts in log space, $\bm\ell=g(\operatorname{vec}(\bm H_{\mathrm c}))\in\mathbb R^5$, where $\operatorname{vec}$ concatenates all six token states. For aggregation, we recover nonnegative counts as $\hat n_k=\max\{0,\exp(\ell_k)-1\}$. Let $\mathcal D$ index categories with individually resolved counts $n_k$, and let $\mathcal G$ contain groups $G\subseteq\{1,\ldots,5\}$ with annotated combined counts $n_G$. We optimize
\begin{equation}
\begin{aligned}
\mathcal L_{\mathrm{count}}
={}&\sum_{k\in\mathcal D}
\operatorname{SmoothL1}\bigl(\ell_k,\log(1+n_k)\bigr)\\
&+\sum_{G\in\mathcal G}
\operatorname{SmoothL1}\!\left(
\log\!\left(1+\sum_{k\in G}\hat n_k\right),
\log(1+n_G)\right).
\end{aligned}
\label{eq:count_loss}
\end{equation}

Merged fine labels supervise individual target counts, whereas coarse labels supervise their combined abundance. For example, a coarse inflammatory annotation aligns both inflammatory token representations while supervising the sum of lymphocyte and other-inflammatory counts.

The frozen teachers and visual-supervision modules are used only during SFT. The count head supervises image-wide abundance; the language model learns to generate regional measurements through response-text supervision. At inference, it generates visual tokens, observations, and answers autoregressively. Non-cellular training samples use only the \texttt{<uni>} group before answer generation.

\subsection{Supervised Fine-Tuning}
\label{sec:sft}
The visual tokens must first acquire useful image representations, then be generated as part of a response and used to support an answer. We organize SFT into three stages, as shown in Figure \ref{fig2}. In \emph{Perceive}, the image, question, and visual tokens are supplied in the user prompt. Visual supervision shapes the token states while the model learns to answer from this context. In \emph{Generate}, the model predicts the visual-token block from the image and a feature query, learning to produce the representations that were previously supplied in the prompt. In \emph{Reason}, it generates the complete response: visual tokens in \texttt{<think>}, measurements in \texttt{<observe>}, and reasoning with a final answer in \texttt{<answer>}. This stage trains the representations, reported measurements, and answer within the same generation sequence.

Cross-entropy covers the entire assistant target and masks the user prompt, supervising visual-token generation in stages two and three. Visual losses apply throughout:
\begin{equation}
\mathcal L_{\mathrm{SFT}}
=\mathcal L_{\mathrm{LM}}
+\lambda_{\mathrm u}\mathcal L_{\mathrm{uni}}
+\lambda_{\mathrm a}\mathcal L_{\mathrm{align}}
+\lambda_{\mathrm c}\mathcal L_{\mathrm{count}},
\label{eq:sft_objective}
\end{equation}
where $\mathcal L_{\mathrm{LM}}$ is assistant-token cross-entropy and the $\lambda$ coefficients weight the three visual losses. A visual loss is zero when its token group or supervision is unavailable. Training updates LoRA parameters, visual-token parameters, and visual-supervision modules. For quantitative examples, programs derive counts and reference answers from nucleus annotations; a language model supplies descriptions and reasoning, checked against those values. Data, training details, and response templates appear in Appendices~\ref{app:data}, \ref{app:implementation}, and~\ref{app:training_prompts}.


\subsection{Reinforcement Learning for Answer--Observation Consistency}
\label{sec:rl}

SFT provides consistent reference responses, but inference requires reasoning from the model's own estimated counts. An answer-only reward can accept a correct answer that contradicts those counts. We therefore train the model on its generated responses, rewarding both answer correctness and agreement with its observations.

\paragraph{Answer correctness and count consistency.}
For question $u$, let $Q$ denote reference counts and $a^\star=f_u(Q)$ the complete reference answer. A response reports counts $C$ and answer $\hat a$. Let $g_u$ denote the count-based task check and $\pi_u$ the corresponding answer projection. We define
\begin{equation}
\begin{aligned}
\mathrm{Acc}&=\mathbf{1}[\hat a=a^\star],\qquad
\mathrm{Con}=\mathbf{1}[g_u(C)=\pi_u(\hat a)],\\
R&=V_s\bigl(\mathrm{Acc}+\beta\,\mathrm{Con}\bigr)-\gamma(1-V_s).
\end{aligned}
\label{eq:rl_reward}
\end{equation}
Acc checks the complete answer, whereas Con compares the count-derived decision with the corresponding part of the model's own answer. For multi-step composition, Con checks region selection only. We set $\mathrm{Con}=0$ when either side is undeterminable. The coefficients $\beta$ and $\gamma$ weight the consistency bonus and invalid-response penalty. The structural indicator $V_s$ requires parseable observation JSON, nonnegative counts and a parseable final answer. Answer correctness and consistency are evaluated with deterministic rules.

\paragraph{Policy optimization.}
Starting from ASPECT-SFT, we optimize the policy with GRPO \citep{shao2024deepseekmath}. We sample a group of responses for each prompt and normalize their rewards within the group to obtain relative advantages for the clipped policy update. LoRA parameters in both the visual and language components are updated. Appendices~\ref{app:rl_data} and~\ref{app:rl_implementation} detail the RL data and optimization settings.

\begin{figure*}[t!]
  \includegraphics[width=1.\textwidth]{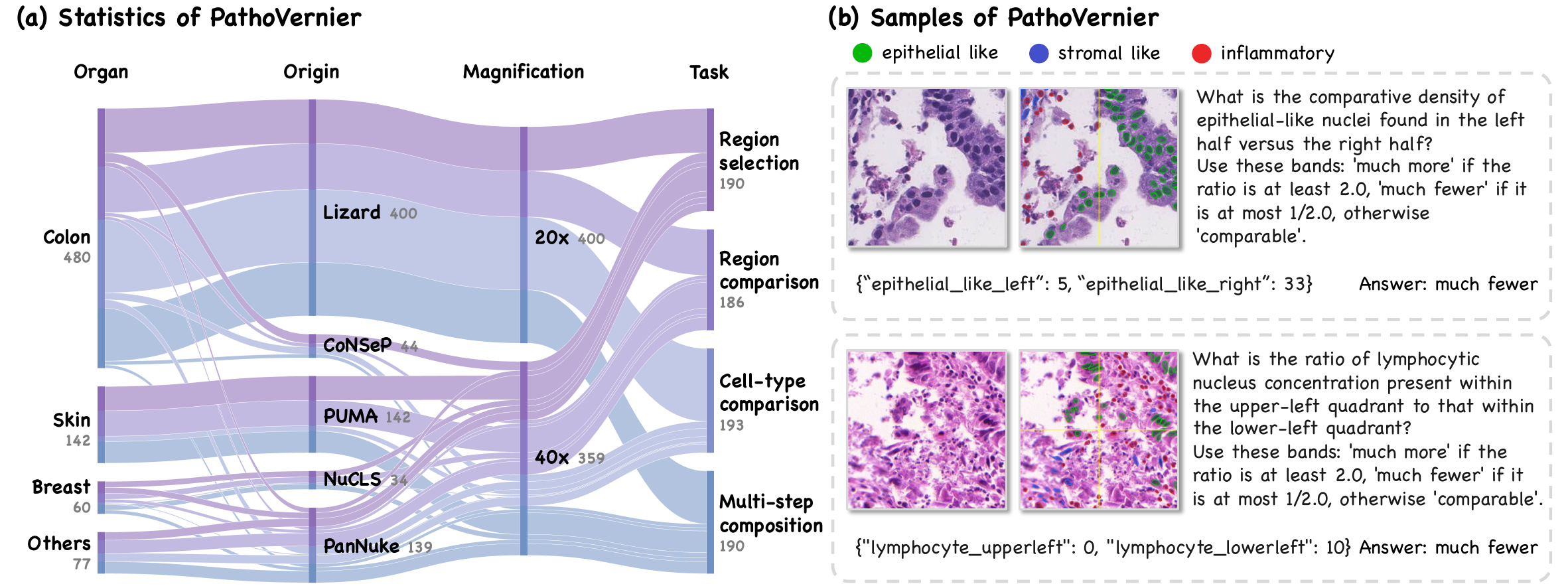}
  \caption{PathoVernier overview. (a) Distribution of 759 questions across organs, sources, magnifications, and tasks; flow widths represent question counts. (b) Example questions with reference counts.}
  \label{fig3}
\end{figure*}

\subsection{PathoVernier Benchmark}
\label{sec:vernier}
\paragraph{Data statistics and task coverage.}
PathoVernier evaluates whether pathology VLMs can answer questions based on accurate cell measurements. As shown in Figure \ref{fig3}(a), it contains 759 expert-reviewed questions across different organs, with images acquired at approximately $20\times$ and $40\times$ magnification. The images originate from five datasets with nucleus-level annotations: Lizard~\citep{graham2021lizard}, PUMA~\citep{schuiveling2025novel}, PanNuke~\citep{gamper2020pannuke}, CoNSeP~\citep{graham2019hover}, and NuCLS~\citep{amgad2022nucls}. The questions are approximately balanced across four tasks. \emph{Region selection} identifies the region with the highest density of a specified cell type. \emph{Region comparison} compares one cell type across two regions, whereas \emph{cell-type comparison} compares two types within the same image. \emph{Multi-step composition analysis} requires selecting the densest region and then determining a cell proportion category within it. These tasks assess cell recognition, spatial localization, and quantitative reasoning, with reference measurements accompanying every answer.

\paragraph{Question construction and annotation.}
We filter source images for task suitability and balance while keeping them disjoint from training images. We then map the original nucleus labels to the cell categories used in the questions and compute reference counts from the nucleus annotations. For a region $\mathcal R$ and a cell category $c$, the annotated count is
\begin{equation}
q_{\mathcal R,c}
=\sum_{j\in\mathcal J_c}\mathbf{1}[\mathbf p_j\in\mathcal R],
\label{eq:annotation_count}
\end{equation}
where $\mathcal J_c$ indexes retained nuclei belonging to category $c$, including its constituent classes for aggregate categories, and $\mathbf p_j$ is nucleus $j$'s centroid.
Four templates convert the required counts $Q$ and rule-derived answers $a^\star=f_u(Q)$ into 786 candidate questions. GPT-5.6-sol \citep{openai2026gpt56sol} rewrites their wording while preserving numerical rules and answers. Experts review each question against its image and annotation overlay, retaining 759 valid items (Figure \ref{fig3}(b)). Appendices~\ref{app:benchmark_construction} and~\ref{app:question_rewriting} detail filtering, rewriting, and expert review.


\paragraph{Evaluation metrics.}
We report answer accuracy $\mathrm{Acc}=\mathbb E[\mathbf 1[\hat a=a^\star]]$ and two measures of the reported observations. All expectations denote empirical averages over eligible responses. Let $\mathcal K$ index a question's required counts, $q_i$ be its annotated count, and $\operatorname{dom}(C)$ index reported counts $C_i$. The fraction of required counts within tolerance is
\begin{equation}
S(C,Q)=\frac{1}{|\mathcal K|}\sum_{i\in\mathcal K\cap\operatorname{dom}(C)}\mathbf 1\bigl[|C_i-q_i|\leq\tau_i\bigr],\qquad\tau_i=\max(1,0.1q_i).
\label{eq:count_score}
\end{equation}
Each count contributes equally; missing counts contribute zero.
We adapt coherent accuracy (CA) from VPRM \citep{pronesti2026beyond}
to compare a count-derived task decision with its reference:
\begin{equation}
\mathrm{CA}
=\mathbb{E}\!\left[
\mathbf{1}[g_u(C)=\pi_u(a^\star)]
\,\middle|\,
g_u(C)\text{ is defined}
\right].
\end{equation}
Here, $g_u$ and $\pi_u$ denote the task check and answer projection defined in Appendix~\ref{app:metric_protocols}. Additionally, right answer, wrong reason (RAWR) measures the fraction of required
counts outside tolerance among correct responses with complete measurements:
\begin{equation}
\mathrm{RAWR}=1-\mathbb E\!\left[S(C,Q)\,\middle|\,\hat a=a^\star,\;\mathcal K\subseteq\operatorname{dom}(C)\right].
\label{eq:rawr}
\end{equation}
Lower RAWR indicates fewer out-of-tolerance counts. Appendix~\ref{app:metric_protocols} specifies eligibility for both conditional metrics.

\section{Experiments}

\begin{table*}[t!]
\renewcommand{\arraystretch}{1.1}
\renewcommand{\tabcolsep}{12pt}
\centering
\caption{Comparison on PathoVernier and three external pathology benchmarks. ASPECT leads on all PathoVernier metrics and improves over Qwen3-VL-8B on all external benchmarks. \textbf{Bold} and \underline{underlined} values denote the best and second-best results, respectively. Dashes indicate CA coverage below 30\%; Table~\ref{tab:metric_coverage} reports coverage and unconditional Count Acc.}
\resizebox{\textwidth}{!}
{\begin{tabular}{l|cccccc}
\toprule[1pt]
\multirow{2}{*}{\textbf{Models}} & \multicolumn{3}{c}{\textbf{PathoVernier}} & \textbf{PathCLS} & \textbf{PathVQA} & \textbf{Quilt-VQA} \\ \cline{2-7} 
                       & \textbf{Acc$\uparrow$}       & \textbf{CA$\uparrow$}        & \textbf{RAWR$\downarrow$}     & \textbf{Acc$\uparrow$}     & \textbf{Acc$\uparrow$}     & \textbf{Acc$\uparrow$}      \\ \midrule[0.7pt]
\multicolumn{7}{c}{\textit{Closed-source Models}}                                                 \\ \midrule[0.7pt]
GPT-5.5 \citep{openai2026gpt55}               & 0.59      & 0.65      & 0.76     & \underline{0.55}    & 0.69    & \textbf{0.69}     \\
Gemini-3.1-Pro \citep{google2026gemini31pro}        & \underline{0.63}      & \underline{0.71}      & \underline{0.68}     & \textbf{0.67}    & \textbf{0.78}    & 0.65     \\ \midrule[0.7pt]
\multicolumn{7}{c}{\textit{Open-source Models}}                                                   \\ \midrule[0.7pt]
Qwen3-VL-8B \citep{bai2025qwen3}              & 0.37      & 0.53      & 0.85     & 0.37    & 0.67    & 0.58     \\
Qwen3-VL-32B \citep{bai2025qwen3}              & 0.44      & 0.50      & 0.82     & 0.45    & 0.65    & 0.62     \\
InternVL3-8B \citep{zhu2025internvl3}             & 0.30      & 0.38      & 0.86     & 0.39    & 0.69    & 0.64     \\
InternVL3-38B \citep{zhu2025internvl3}             & 0.45      & 0.45      & 0.83     & 0.41    & 0.71    & 0.64     \\ \midrule[0.7pt]
\multicolumn{7}{c}{\textit{Medical and Pathology-tuned Models}}                                               \\ \midrule[0.7pt]
Quilt-LLaVA-7B \citep{seyfioglu2024quilt}           & 0.12      & -         & -        & 0.24    & 0.54    & 0.64     \\
PathGen-LLaVA-13B \citep{sun2025pathgen}         & 0.12      & -         & -        & 0.52    & 0.62    & 0.66     \\
HuatuoGPT-Vision-7B \citep{chen2024towardsh}      & 0.30      & -         & -        & 0.34    & 0.64    & 0.55     \\
Patho-R1-7B \citep{zhang2026patho}              & 0.22      & -         & -        & 0.44    & 0.56    & \underline{0.68}     \\
\rowcolor{blue!10}\textbf{ASPECT-8B(ours)}           & \textbf{0.75}      & \textbf{0.81}      & \textbf{0.49}     & 0.54    & \underline{0.74}    & 0.67     \\ 
\rowcolor{blue!10}$\Delta$ (\textit{vs base model Qwen3-VL-8B})          & \textcolor{green!60!black}{\textbf{+0.38}}      & \textcolor{green!60!black}{\textbf{+0.28}}      & \textcolor{green!60!black}{\textbf{-0.36}}     & \textcolor{green!60!black}{\textbf{+0.17}}    & \textcolor{green!60!black}{\textbf{+0.07}}    & \textcolor{green!60!black}{\textbf{+0.09}}     \\ 
\bottomrule[1pt]
\end{tabular}
}
\label{table1}
\end{table*}

\paragraph{Performance on PathoVernier.}
Table~\ref{table1} shows that ASPECT leads on all three PathoVernier metrics, with relative gains of approximately 19.2\% in Acc and 12.9\% in CA, and a 28.1\% reduction in RAWR over Gemini-3.1-Pro, the strongest baseline. ASPECT thus improves both task decisions and the quantitative observations underlying them. The reduction in RAWR directly addresses our motivating observation: even among correct responses, ASPECT's reported counts more often match the annotated measurements. In contrast, all medical- and pathology-tuned baselines fall below Qwen3-VL-8B in PathoVernier accuracy and have CA coverage below 30\%, so we omit their CA and RAWR. Notably, PathGen-LLaVA and Patho-R1 outperform Qwen3-VL-8B on PathCLS yet underperform it on PathoVernier. This ranking reversal shows that conventional pathology question-answering performance does not guarantee reliable quantitative image understanding. Appendices~\ref{app:task_results}--\ref{app:tolerance_results} report task- and source-level results, confidence intervals, and tolerance sensitivity.

\begin{figure*}[t!]
  \includegraphics[width=1.\textwidth]{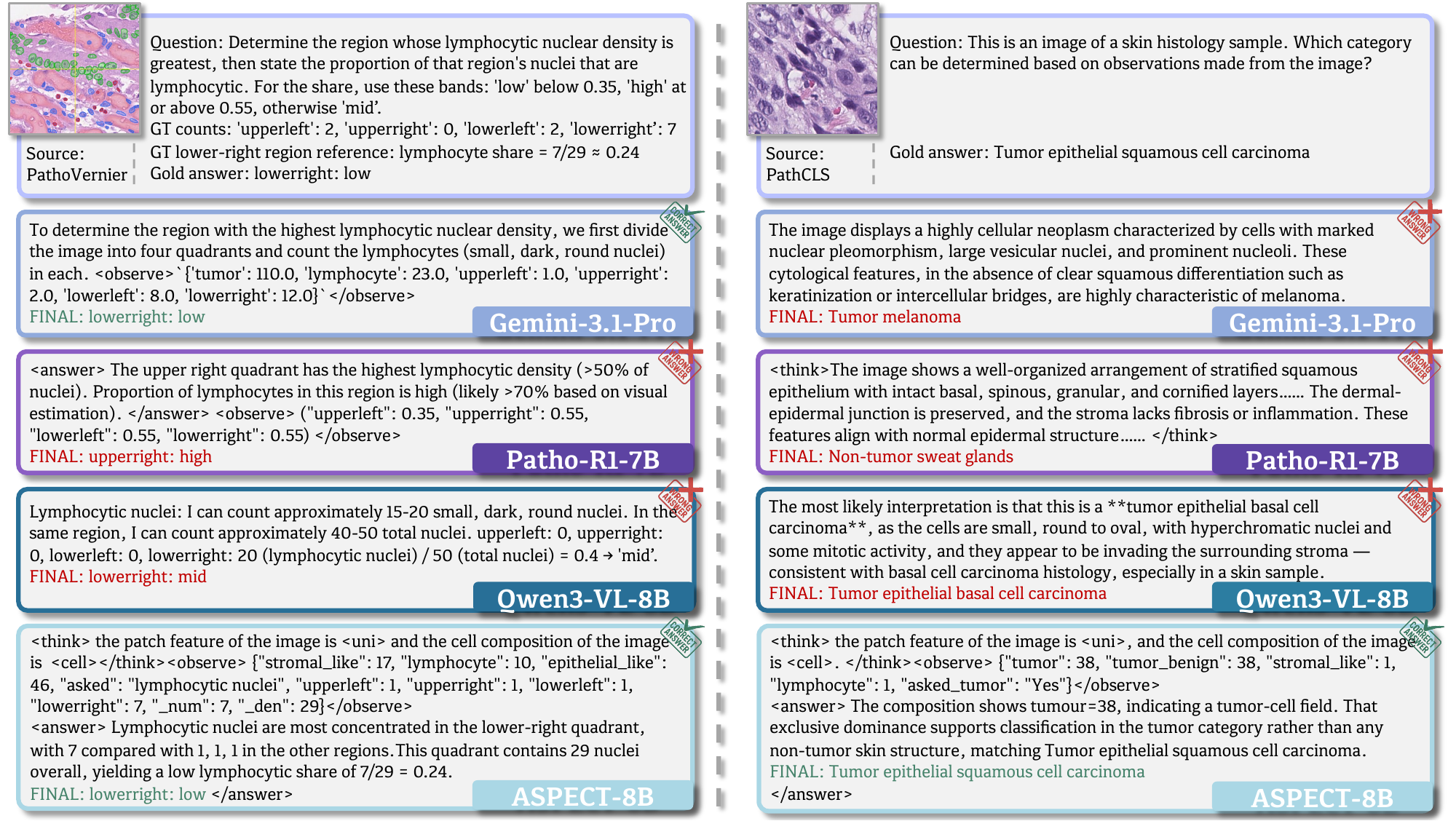}
  \caption{Qualitative comparison on PathoVernier (left) and PathCLS (right). Responses are condensed and reformatted to highlight key observations and predictions. ASPECT reports more accurate regional counts on PathoVernier and predicts the reference class on PathCLS.}
  \label{fig4}
\end{figure*}

\paragraph{Performance on external pathology benchmarks.}
ASPECT also improves over its Qwen3-VL-8B backbone on all three external pathology benchmarks, with relative accuracy gains of approximately 45.9\% on PathCLS~\citep{sun2024pathmmu}, 10.4\% on PathVQA~\citep{he2020pathvqa}, and 15.5\% on the pathology-image subset of Quilt-VQA~\citep{seyfioglu2024quilt}. The gains on classification and question answering show that training for cellular composition also benefits pathology tasks that do not explicitly request measurements.

\paragraph{Qualitative comparison.}
Figure~\ref{fig4} summarizes predictions from four models. On PathoVernier, Gemini and ASPECT give the same correct answer, but Gemini overestimates the lower-right lymphocyte count (12 versus 7). ASPECT's regional counts fall within tolerance, and its reported proportion, $7/29\approx0.24$, correctly supports the \texttt{low} category. Qwen3-VL-8B selects the correct region but misjudges the proportion; Patho-R1 errs on both steps. This contrast shows how answer accuracy can conceal measurement errors. On PathCLS, ASPECT predicts the reference class while all three baselines misclassify the image, illustrating a classification example beyond composition analysis. Full responses for these and additional cases are provided in Appendix~\ref{app:case_studies}.

\begin{table*}[t!]
\renewcommand{\arraystretch}{1.1}
\renewcommand{\tabcolsep}{8pt}
\centering
\caption{Ablations of visual supervision and the SFT curriculum on PathoVernier. All variants are evaluated before RL with matched total update budgets.}
\resizebox{0.8\textwidth}{!}
{
\begin{tabular}{l|cccc}
\toprule[1pt]
\textbf{Configuration}                     & \textbf{Acc$\uparrow$} & \textbf{CA$\uparrow$} & \textbf{RAWR$\downarrow$} & \textbf{Count Acc$\uparrow$} \\ \midrule[0.7pt]
ASPECT-SFT                           & 0.70           & 0.80          & 0.53            & 0.44                 \\
w/o visual supervision                & 0.64           & 0.77          & 0.59            & 0.32                 \\
w/o pathology feature reconstruction & 0.68           & 0.75          & 0.56            & 0.27                 \\
w/o cell feature alignment           & 0.67           & 0.77          & 0.54            & 0.29                 \\
w/o count supervision                & 0.72           & 0.78          & 0.57            & 0.36                 \\
Direct Reason training (Stage 3)     & 0.62           & 0.74          & 0.56            & 0.31                 \\ \bottomrule[1pt]
\end{tabular}
}
\label{table2}
\end{table*}

\begin{table*}[t!]
\renewcommand{\arraystretch}{1.1}
\renewcommand{\tabcolsep}{8pt}
\centering
\caption{RL reward ablations on PathoVernier. The listed rewards are the internal term $r$; all RL variants retain the structural validity gate and the invalid-response penalty.}
\resizebox{\textwidth}{!}
{
\begin{tabular}{ll|ccccc}
\toprule[1pt]
\multicolumn{1}{c}{\textbf{Configuration}} & \textbf{RL Reward} & \textbf{Acc$\uparrow$} & \textbf{CA$\uparrow$} & \textbf{RAWR$\downarrow$} & \textbf{Count Acc$\uparrow$} & \textbf{Consistency$\uparrow$} \\ \midrule[0.7pt]
ASPECT-SFT                                 & N/A                & 0.70           & 0.80          & 0.53          & 0.44                 & 0.94                   \\
Answer only                                & Acc                & 0.72           & 0.79          & 0.55          & 0.43                 & 0.92                   \\
Consistency only                           & 0.5 Con            & 0.69           & 0.76          & 0.51          & 0.41                 & 0.98                   \\
Answer + count score                       & Acc + 0.5 S    & 0.74           & 0.80          & 0.50          & 0.45                 & 0.96                   \\
ASPECT-8B                                  & Acc + 0.5 Con      & 0.75           & 0.81          & 0.49          & 0.47                 & 0.99                   \\ \bottomrule[1pt]
\end{tabular}
}
\label{table3}
\end{table*}

\begin{figure*}[t!]
  \setlength{\abovecaptionskip}{0pt}
  \includegraphics[width=1.\textwidth]
  {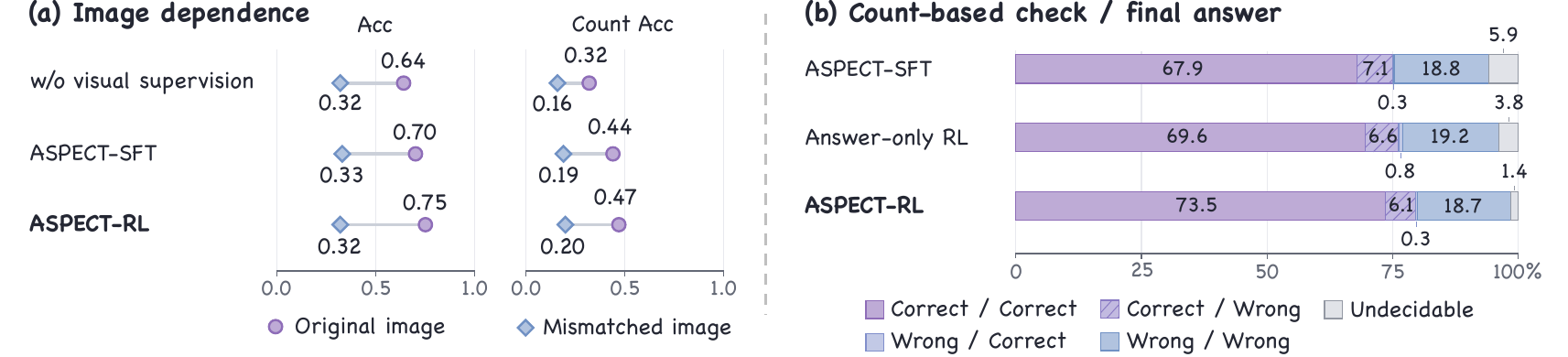}
  \caption{Image dependence and task correctness on PathoVernier. (a) Performance with original and mismatched images. (b) Joint outcomes of the count-based check and final-answer correctness.}
  \label{fig5}
\end{figure*}

\paragraph{Ablation on visual supervision and training curriculum.}
Table \ref{table2} ablates visual supervision and the SFT curriculum on PathoVernier. Count Acc averages $S(C,Q)$ over all questions, while RAWR measures count errors only among correct answers with complete measurements. ASPECT-SFT achieves 0.44 Count Acc, compared with 0.296 for Gemini-3.1-Pro and 0.068 for Qwen3-VL-8B. Removing pathology feature reconstruction or cell alignment reduces Count Acc to 0.27 or 0.29. Removing count supervision slightly improves Acc but lowers Count Acc to 0.36 and increases RAWR to 0.57. Thus, reported measurements can support the correct conclusion while remaining inaccurate, reinforcing the need to evaluate observations alongside answers. The staged SFT configuration also improves answer and count accuracy over QA-only Direct Reason training.

\paragraph{Ablation on reinforcement learning rewards.}
Table \ref{table3} compares reward variants from the same SFT checkpoint, retaining $R=V_s r-0.1(1-V_s)$ and varying only $r$, with count score $S(C,Q)$. Consistency averages $\mathrm{Con}$ over responses with determinable count-derived decisions and answer components, measuring agreement with the model's own decision; CA instead uses the reference decision. Answer-only RL improves Acc while degrading count accuracy and consistency. Conversely, consistency-only RL reaches 0.98 agreement without improving answer or count accuracy over SFT. ASPECT combines these complementary signals and achieves the highest Acc and Count Acc and the lowest RAWR among the tested rewards.

\paragraph{Image dependence and task correctness.}
Figure \ref{fig5}(a) shows that mismatching images largely removes the Acc gains and reduces the Count Acc gains, indicating that ASPECT’s improvements depend on relevant visual evidence. Figure \ref{fig5}(b) further examines a gold-anchored count-based check alongside final-answer correctness. ASPECT-RL passes both checks on 73.5\% of questions, versus 67.9\% for SFT and 69.6\% for answer-only RL. For multi-step composition, the count-based check assesses the reference region, whereas final-answer accuracy requires both the region and proportion band (Appendix~\ref{app:ablation_protocols}).

\section{Conclusion}
We presented ASPECT and PathoVernier to connect fine-grained visual perception with pathology reasoning. ASPECT supervises intermediate visual tokens for pathology appearance and cell abundance, then learns to generate and use quantitative observations through staged SFT and RL. PathoVernier evaluates the resulting answers together with their supporting measurements. ASPECT improves both answer accuracy and count fidelity, with further gains over its backbone on three external pathology benchmarks. Ablations and image-mismatch experiments show the value of visual supervision and the dependence of these gains on relevant images. Together, these results demonstrate the value of treating quantitative observations as explicit targets for pathology VLM training and evaluation.

\subsection*{Reproducibility Statement}
Code: \url{https://github.com/ChyaZhang/ASPECT}\\
Model: \url{https://huggingface.co/Mikezcy/ASPECT-8B}\\
Benchmark: \url{https://huggingface.co/datasets/Mikezcy/PathoVernier}



\bibliography{iclr2027_conference}
\bibliographystyle{iclr2027_conference}

\clearpage
\appendix
\setcounter{figure}{0} 
\renewcommand{\thefigure}{\thesection.\arabic{figure}}
\setcounter{table}{0}
\renewcommand{\thetable}{\thesection.\arabic{table}}
\renewcommand{\theHtable}{appendix.\arabic{table}}
\renewcommand{\theHfigure}{appendix.\arabic{figure}}
\section*{Appendix}
\section*{Contents of the Appendix}

\begin{itemize}
    \setlength{\itemsep}{3pt}

    \item \hyperref[app:data]{%
        \ref*{app:data}. Data Sources and Construction}

    \item \hyperref[app:implementation]{%
        \ref*{app:implementation}. Implementation Details and Algorithms}

    \item \hyperref[app:evaluation]{%
        \ref*{app:evaluation}. Evaluation Protocols and Metric Definitions}

    \item \hyperref[app:prompts]{%
        \ref*{app:prompts}. Prompts and Response Formats}

    \item \hyperref[app:additional_results]{%
        \ref*{app:additional_results}. Additional Quantitative Results}

    \item \hyperref[app:case_studies]{%
        \ref*{app:case_studies}. Complete Case Studies and Failure Analysis}

    \item \hyperref[app:limitations]{%
        \ref*{app:limitations}. Limitations and Future Work}

\end{itemize}

\section{Data Sources and Construction}
\label{app:data}

\subsection{Source Datasets and Data Splits}
\label{app:data_sources}

We use seven public datasets for training and benchmark construction (Table~\ref{tab:data_sources}). Lizard \citep{graham2021lizard}, PUMA \citep{schuiveling2025novel}, PanNuke \citep{gamper2020pannuke}, CoNSeP \citep{graham2019hover}, and NuCLS \citep{amgad2022nucls} provide nucleus-level
annotations for quantitative supervision and PathoVernier construction. The first four provide instance masks and cell labels, whereas NuCLS provides nucleus locations and class annotations for the subset used in this work.
NCT-CRC-HE-100K \citep{kather2018nctcrc} and the H\&E subset of the PathVQA \citep{he2020pathvqa} training split add tissue-classification and pathology question-answering examples during SFT.
For NCT-CRC-HE-100K, cellular supervision combines detected nuclei with patch-level tissue labels. For PathVQA, it uses predicted nucleus locations and native cell-type predictions. Neither source provides reference nucleus-level annotations.

Data are partitioned according to the available source identifiers: source images for Lizard, source WSIs and their corresponding ROIs for PUMA, individual patches for PanNuke, and TCGA patient--slide identifiers for NuCLS. CoNSeP follows its official training and test partition. NCT-CRC-HE-100K is sampled from its official training set with balanced tissue classes, while PathVQA uses only its official training split after H\&E filtering. SFT, RL, validation, and PathoVernier use mutually disjoint image sets. The SFT collection contains 19,422 visual-supervision examples and 13,095 question-answer pairs. RL uses 1,372 questions from 1,180 unique images, with 93 validation questions used for checkpoint selection. PathoVernier contains 759 questions from 553 unique patches.

\paragraph{Data access and release.}
The PathoVernier release will include questions, reference counts, split indices, and scripts for reconstructing the benchmark from the original datasets. Images will be obtained directly from the respective providers rather than redistributed with the benchmark. The release documentation will retain source attribution and specify the applicable licenses and access terms for source-derived materials.

\begin{table*}[h!]
\renewcommand{\arraystretch}{1.1}
\renewcommand{\tabcolsep}{6pt}
\centering
\caption{Data sources and sample counts. SFT examples are separated into visual-supervision and question-answer examples. RL, validation, and PathoVernier columns report question counts. Multiple examples may share an image.}
\resizebox{\textwidth}{!}
{
\begin{tabular}{lccccc}
\toprule[1pt]
\multirow{2}{*}{Source} & \multicolumn{2}{c}{SFT} & \multirow{2}{*}{RL} & \multirow{2}{*}{Validation} & \multirow{2}{*}{PathoVernier} \\ \cline{2-3}
                        & Visual      & VQA        &                     &                             &                               \\ \midrule[0.7pt]
Lizard \citep{graham2021lizard}                  & 2,520      & 2,944      & 559                 & 40                          & 400                           \\
PUMA \citep{schuiveling2025novel}                   & 2,745      & 1,010      & 219                 & 5                           & 142                           \\
PanNuke \citep{gamper2020pannuke}                & 6,780      & 2,565      & 567                 & 20                          & 139                           \\
CoNSeP \citep{graham2019hover}                 & 196        & 133        & 27                  & 2                           & 44                            \\
NuCLS \citep{amgad2022nucls}                  & 854        & 77         & 0                   & 26                          & 34                            \\
NCT-CRC-HE-100K \citep{kather2018nctcrc}        & 5,938      & 5,386      & 0                   & 0                           & 0                             \\
PathVQA \citep{he2020pathvqa}                & 389        & 980        & 0                   & 0                           & 0                             \\ \midrule[0.7pt]
Total                   & 19,422     & 13,095     & 1,372               & 93                          & 759                           \\ \bottomrule[1pt]
\end{tabular}
}
\label{tab:data_sources}
\end{table*}

\subsection{Cell Category Harmonization}
\label{app:category_mapping}

With guidance from pathologists, we harmonize source labels into five categories: tumor cells, normal epithelial cells, stromal-like cells, lymphocytes, and other inflammatory cells (Table~\ref{tab:category_mapping}). Fine-grained labels within a category are merged. Coarse labels retain their aggregate meaning: Lizard epithelial annotations remain epithelial-like, while PanNuke and CoNSeP inflammatory annotations remain inflammatory. For feature alignment, these annotations contribute to both corresponding category targets; count supervision applies only to their combined count.

Background and excluded labels are removed before constructing supervision: Dead in PanNuke, Other/Miscellaneous in CoNSeP, apoptotic nuclei in PUMA, and unlabeled nuclei or apoptotic bodies in NuCLS. These instances contribute neither to category counts nor to the total count or the sixth cell-token target. The sixth token summarizes all retained, mappable instances, each included once.

For sources without nucleus-level annotations, we use two supervision paths (Table~\ref{tab:external_cell_supervision}). In NCT-CRC-HE-100K, CellViT-SAM-H detects nuclei, and all detected nuclei in a patch inherit the category mapped from its tissue label. Adenocarcinoma epithelium maps to tumor, normal colon mucosa to normal epithelial, stroma and smooth muscle to stromal-like, and lymphocytes to lymphocyte. These assignments provide weak cell-type labels. Adipose, debris, background, and mucus patches receive pathology feature reconstruction without cell-specific supervision. In PathVQA, native CellViT PanNuke predictions follow the PanNuke mapping in Table~\ref{tab:category_mapping}, including aggregate inflammatory supervision and exclusion of Dead and background predictions.

Benchmark questions use tumor, stromal-like, lymphocyte, epithelial-like, and inflammatory counts. Epithelial-like combines tumor and normal epithelial cells; inflammatory combines lymphocytes and other inflammatory cells. These quantities are obtained by summing resolved categories or retaining the corresponding aggregate annotation. Questions respect source-label granularity: Lizard does not support tumor-specific questions, while PanNuke and CoNSeP do not support lymphocyte-specific questions. Normal epithelial and other inflammatory counts are not queried individually. Cell-type comparisons exclude category pairs with a subset relationship, such as tumor versus epithelial-like.

\begin{table*}[h!]
\renewcommand{\arraystretch}{1.1}
\setlength{\tabcolsep}{6pt}
\centering
\caption{Source-label mappings for cellular supervision. Starred entries retain aggregate labels: feature alignment uses both listed targets, whereas count supervision uses their sum. NuCLS uses these mappings for count supervision only.}
\resizebox{\textwidth}{!}{
\begin{tabular}{lll}
\toprule[1pt]
Source & Original label(s) & Target category/categories \\
\midrule[0.7pt]
\multirow{4}{*}{Lizard}
& Epithelial & Tumor; normal epithelial$^\ast$ \\
& Connective tissue & Stromal-like \\
& Lymphocyte & Lymphocyte \\
& Plasma, neutrophil, eosinophil & Other inflammatory \\
\midrule[0.7pt]
\multirow{5}{*}{PUMA}
& Tumor & Tumor \\
& Epithelium & Normal epithelial \\
& Stroma, endothelium & Stromal-like \\
& Lymphocyte & Lymphocyte \\
& Plasma cell, neutrophil, histiocyte, melanophage & Other inflammatory \\
\midrule[0.7pt]
\multirow{4}{*}{PanNuke}
& Neoplastic & Tumor \\
& Epithelial & Normal epithelial \\
& Connective/soft tissue & Stromal-like \\
& Inflammatory & Lymphocyte; other inflammatory$^\ast$ \\
\midrule[0.7pt]
\multirow{4}{*}{CoNSeP}
& Dysplastic/malignant epithelial & Tumor \\
& Healthy epithelial & Normal epithelial \\
& Fibroblast, muscle, endothelial & Stromal-like \\
& Inflammatory & Lymphocyte; other inflammatory$^\ast$ \\
\midrule[0.7pt]
\multirow{5}{*}{NuCLS}
& Tumor, mitotic figure & Tumor \\
& Ductal epithelium, myoepithelium & Normal epithelial \\
& Fibroblast, vascular endothelium & Stromal-like \\
& Lymphocyte & Lymphocyte \\
& Plasma cell, macrophage, neutrophil, eosinophil & Other inflammatory \\
\bottomrule[1pt]
\end{tabular}
}
\label{tab:category_mapping}
\end{table*}

\begin{table*}[h!]
\renewcommand{\arraystretch}{1.1}
\setlength{\tabcolsep}{6pt}
\centering
\caption{Cellular supervision for sources without nucleus-level reference annotations. CRC cell types inherit patch-level tissue labels; PathVQA cell types come from native nucleus predictions.}
\resizebox{\textwidth}{!}{
\begin{tabular}{llll}
\toprule[1pt]
Source & Label source & Feature-alignment targets & Count supervision \\
\midrule[0.7pt]
NCT-CRC-HE-100K
& \shortstack[l]{Patch-level\\tissue labels}
& \shortstack[l]{Mapped patch category\\for all detected nuclei}
& \shortstack[l]{Detected nucleus count\\assigned to that category} \\
\midrule[0.7pt]
PathVQA
& \shortstack[l]{Native CellViT\\PanNuke predictions}
& \shortstack[l]{Mapped nucleus categories;\\shared inflammatory targets}
& \shortstack[l]{Predicted category counts;\\aggregate inflammatory count} \\
\bottomrule[1pt]
\end{tabular}
}
\label{tab:external_cell_supervision}
\end{table*}

\subsection{SFT Data and Teacher Targets}
\label{app:sft_data}
The SFT collection contains 19,422 visual-supervision records and 13,095 question-answer examples. These records are rendered into stage-specific input--target pairs: the visual-supervision records are used in Generate, while QA records are used in all three stages. The question-answer examples comprise 6,729 programmatically constructed quantitative questions from the five nucleus-annotated datasets, 5,386 classification and related questions from NCT-CRC-HE-100K, and 980 closed-ended yes/no questions from the H\&E subset of PathVQA. GPT-5.6-sol generates descriptive prose and reasoning from the image and source-specific reference information. In a subsequent text-only pass, the same model rewrites the \texttt{\textless answer\textgreater} reasoning notes while preserving their numerical values and final answers. PathVQA observations are qualitative and contain no count JSON. Automatic checks enforce agreement between the generated prose, reference counts, and answers. Failed generations are retried or discarded; rewrites that alter numerical values or final answers revert to templates.

Visual teacher targets are extracted offline and remain fixed during training. UNI provides image features for pathology feature reconstruction. For Lizard, PUMA, PanNuke, and CoNSeP, CellViT-SAM-H features are mean-pooled within ground-truth nucleus masks to obtain instance embeddings, then normalized and aggregated into six cell-token targets as described in Section \ref{sec:visual_supervision}. NuCLS supplies UNI features and category counts derived from point annotations, but does not contribute to cell feature alignment. NCT-CRC-HE-100K and PathVQA use the detected instances and category assignments described in Appendix~\ref{app:category_mapping}. PathVQA patches with fewer than three retained detections receive UNI supervision only.

\subsection{RL Data Construction}
\label{app:rl_data}
The RL collection contains 1,372 questions from 1,180 unique images drawn from Lizard, PUMA, PanNuke, and CoNSeP. Reference counts are derived from nucleus annotations, and task rules determine the reference answers. RL images are disjoint from those used for SFT and PathoVernier. The collection covers eight skills, including regional selection, composition analysis, and additional counting and comparison tasks (Table~\ref{tab:rl_data}).

We select questions by skill category using the SFT model's rollout accuracy to identify skills with greater room for improvement. We retain the available questions for multi-step composition, region selection, dominant cell type, count band prediction, and count comparison, while limiting the more saturated region-comparison, cell-type-comparison, and yes/no categories. This allocation concentrates training on skills where SFT remains less reliable while preserving task diversity. A 93-question validation set is used for checkpoint selection. Reward and optimization settings are provided in Appendix~\ref{app:rl_implementation}.

\begin{table*}[h!]
\renewcommand{\arraystretch}{1.1}
\setlength{\tabcolsep}{6pt}
\centering
\caption{Skill composition of the RL training collection. Question selection is based on skill-level SFT performance.}
\resizebox{0.7\textwidth}{!}{
\begin{tabular}{lclc}
\toprule[1pt]
Skill & Questions & Skill & Questions \\
\midrule[0.7pt]
Multi-step composition & 388 & Count comparison & 85 \\
Region selection & 407 & Region comparison & 60 \\
Dominant cell type & 208 & Cell-type comparison & 60 \\
Count band prediction & 124 & Yes/no & 40 \\
\midrule[0.7pt]
\multicolumn{3}{l}{Total} & 1,372 \\
\bottomrule[1pt]
\end{tabular}
}
\label{tab:rl_data}
\end{table*}

\subsection{PathoVernier Construction and Expert Review}
\label{app:benchmark_construction}
We construct PathoVernier from eligible H\&E patches in the five nucleus-annotated datasets. After the category filtering described in Appendix~\ref{app:category_mapping}, patches containing fewer than five or at least 300 retained nuclei are excluded. NuCLS contributes only questions whose reference quantities can be computed from point annotations. Candidate questions are sampled with quotas stratified by task, dataset, and answer category to balance task coverage and answer distributions. Four task templates instantiate questions from annotation-derived counts and deterministic decision rules, yielding 786 candidates.

GPT-5.6-sol paraphrases the questions while numerical thresholds and answer options remain fixed. Automatic checks preserve the queried cell types, regions, and relationships; a separate language model checks semantic equivalence against the reference question. Paraphrases that fail these checks are regenerated, with the original template retained when retries are exhausted. An independent implementation recomputes regional assignments and rule-based answers from the annotations and flags discrepancies for manual review. Detailed task rules and rewriting prompts are provided in Appendices~\ref{app:metric_protocols} and~\ref{app:question_rewriting}, respectively.

All 786 candidates are presented through a structured review interface to four board-certified pathologists, with automatically flagged cases prioritized for inspection. Review assesses whether the question matches the image, the reference counts agree with the annotations and visible morphology, and the answer is reliably distinguishable. Particular attention is given to near-tied regional counts and proportions close to category thresholds, especially when small denominators make the answer sensitive to counting discrepancies. Following review, 27 questions are excluded, leaving 759 questions from 553 unique patches with approximately balanced coverage of the four tasks.

\subsection{Detailed Dataset Statistics}
\label{app:data_statistics}

PathoVernier contains 759 questions from 553 unique patches. The four tasks each account for 24.5--25.4\% of the questions, with their source composition detailed in Table~\ref{tab:benchmark_distribution}. We harmonize the 19 source organ labels into 18 categories by merging Colon and Colorectal (Table~\ref{tab:organ_distribution}). Colon, skin, and breast contribute 480, 142, and 60 questions, respectively. The remaining 15 categories contribute 77 questions and are supplied by PanNuke. All distributions in this section count questions, including those sharing the same patch.

\begin{table*}[h!]
\renewcommand{\arraystretch}{1.1}
\setlength{\tabcolsep}{6pt}
\centering
\caption{PathoVernier question counts by source dataset and task.}
\resizebox{\textwidth}{!}{
\begin{tabular}{lccccc}
\toprule[1pt]
Source & Region selection & Region comparison & Cell-type comparison & Multi-step composition & Total \\
\midrule[0.7pt]
Lizard & 82 & 85 & 135 & 98 & 400 \\
PUMA & 45 & 47 & 10 & 40 & 142 \\
PanNuke & 35 & 36 & 33 & 35 & 139 \\
CoNSeP & 17 & 7 & 13 & 7 & 44 \\
NuCLS & 11 & 11 & 2 & 10 & 34 \\
\midrule[0.7pt]
Total & 190 & 186 & 193 & 190 & 759 \\
\bottomrule[1pt]
\end{tabular}
}
\label{tab:benchmark_distribution}
\end{table*}

Each patch contains $256\times256$ pixels at its source resolution. Physical resolution ranges from 0.20 to 0.50~$\mu$m/pixel, corresponding to patch widths of 51.2--128~$\mu$m (Table~\ref{tab:benchmark_scale}). The 400 Lizard questions use images at approximately $20\times$, while the remaining 359 questions use images at approximately $40\times$. These magnification groups therefore also differ in data source and organ composition.

\begin{table*}[h!]
\renewcommand{\arraystretch}{1.1}
\setlength{\tabcolsep}{6pt}
\centering
\caption{PathoVernier question counts across 18 harmonized organ categories. Colon includes the Colorectal label from CoNSeP.}
\resizebox{0.6\textwidth}{!}{
\begin{tabular}{lclc}
\toprule[1pt]
Organ & Questions & Organ & Questions \\
\midrule[0.7pt]
Colon & 480 & Kidney & 4 \\
Skin & 142 & Ovary & 3 \\
Breast & 60 & Liver & 2 \\
Bile duct & 15 & Pancreas & 2 \\
Head and neck & 11 & Prostate & 2 \\
Uterus & 10 & Stomach & 2 \\
Esophagus & 9 & Testis & 2 \\
Cervix & 6 & Thyroid & 2 \\
Lung & 6 & Bladder & 1 \\
\midrule[0.7pt]
\multicolumn{3}{l}{Total} & 759 \\
\bottomrule[1pt]
\end{tabular}
}
\label{tab:organ_distribution}
\end{table*}

\begin{table*}[h!]
\renewcommand{\arraystretch}{1.1}
\setlength{\tabcolsep}{6pt}
\centering
\caption{Image scale by source. Patch width is calculated from the $256\times256$-pixel inputs and source resolution. Magnifications are approximate objective equivalents.}
\resizebox{\textwidth}{!}{
\begin{tabular}{lcccc}
\toprule[1pt]
Source & Resolution ($\mu$m/pixel) & Patch width ($\mu$m) & Approx. magnification & Questions \\
\midrule[0.7pt]
Lizard & 0.50 & 128.0 & $20\times$ & 400 \\
PanNuke, CoNSeP & 0.25 & 64.0 & $40\times$ & 183 \\
PUMA & 0.22 & 56.3 & $40\times$ & 142 \\
NuCLS & 0.20 & 51.2 & $40\times$ & 34 \\
\midrule[0.7pt]
\multicolumn{4}{l}{Total} & 759 \\
\bottomrule[1pt]
\end{tabular}
}
\label{tab:benchmark_scale}
\end{table*}

\section{Implementation Details and Algorithms}
\label{app:implementation}

\subsection{Model and Training Configurations}
\label{app:architecture}

ASPECT uses Qwen3-VL-8B~\citep{bai2025qwen3} as its backbone and applies LoRA~\citep{hu2021lora} to both the visual and language components during SFT and RL. Both stages use AdamW with BF16 training. Table~\ref{tab:training_config} summarizes the principal configurations; additional optimization settings are provided in Appendices~\ref{app:sft_implementation} and~\ref{app:rl_implementation}.

\begin{table*}[h!]
\renewcommand{\arraystretch}{1.1}
\setlength{\tabcolsep}{6pt}
\centering
\caption{Model and training configurations of ASPECT. SFT stage lengths are reported as updates within each stage.}
\resizebox{0.8\textwidth}{!}{
\begin{tabular}{ll}
\toprule[1pt]
Setting & Value \\
\midrule[0.7pt]
\multicolumn{2}{l}{\textit{Model and adaptation}} \\
Backbone & Qwen3-VL-8B \\
Pathology feature tokens & 8 \\
Cell tokens & 6 \\
Visual teachers & UNI; CellViT-SAM-H \\
Teacher updates & Frozen; targets extracted offline \\
LoRA-adapted components & Visual and language components \\
LoRA rank $r$ & 16 \\
LoRA scaling $\alpha$ & 32 \\
Hardware & $4\times$ NVIDIA RTX 6000D \\
\midrule[0.7pt]
\multicolumn{2}{l}{\textit{Supervised fine-tuning}} \\
Total updates & 3,500 \\
Perceive / Generate / Reason updates & 1,000 / 1,000 / 1,500 \\
Effective batch size & 96 \\
LoRA learning rate & $2\times10^{-4}$ \\
Reconstruction and alignment module learning rate & $4 \times 10^{-5}$ \\
New-token embedding learning rate & $4 \times 10^{-5}$ \\
Count head and cell-role embedding learning rate & $2 \times 10^{-4}$ \\
Reconstruction weight $\lambda_{\mathrm{u}}$ & 0.5 \\
Cell-alignment weight $\lambda_{\mathrm{a}}$ & 1.0 \\
Count-supervision weight $\lambda_{\mathrm{c}}$ & 1.0 \\
\midrule[0.7pt]
\multicolumn{2}{l}{\textit{Reinforcement learning}} \\
Initialization & ASPECT-SFT \\
Algorithm & GRPO \\
Learning rate & $1\times10^{-5}$ \\
Responses per question & 8 \\
Sampling temperature & 1.0 \\
Consistency reward coefficient $\beta$ & 0.5 \\
Invalid-response penalty coefficient $\gamma$ & 0.1 \\
\bottomrule[1pt]
\end{tabular}
}
\label{tab:training_config}
\end{table*}

\subsection{Supervised Fine-Tuning}
\label{app:sft_implementation}
SFT proceeds through Perceive, Generate, and Reason for 1,000, 1,000, and 1,500 updates, respectively. Perceive uses QA records, placing visual tokens in the input and supervising the answer block. Generate uses the full pool of visual-supervision and QA records to predict the visual-token block from the image and a feature query. Reason uses QA records to generate the complete \texttt{<think>}, \texttt{<observe>}, and \texttt{<answer>} sequence. All stages use cross-entropy with unit weight over the entire assistant target, masking user-prompt positions. Visual losses apply wherever their supervision is available; NuCLS supplies reconstruction and count supervision without cell feature alignment. Teacher targets remain fixed throughout training.

We use AdamW with a cosine learning-rate schedule, a warmup ratio of 0.05, and weight decay of 0.1. Learning rates and visual-loss weights are listed in Table~\ref{tab:training_config}. A per-device micro-batch size of 1 and gradient accumulation over 24 steps across four GPUs yield an effective batch size of 96. Images are resized to $512\times512$, matching the resolution used for teacher feature extraction. We use random seed 0 and select the checkpoint at update 3,500 as ASPECT-SFT based on the independent development set.

\subsection{Reinforcement Learning}
\label{app:rl_implementation}
RL initializes from ASPECT-SFT and updates LoRA parameters in both the visual and language components using GRPO. Each rollout batch contains 48 questions, with eight responses sampled per question at temperature 1.0. Prompt and response lengths are capped at 2,048 and 512 tokens, respectively. Responses are scored using ~\eqref{eq:rl_reward}, and rewards are normalized within each question's response group to obtain relative advantages. The policy objective uses asymmetric clipping with lower and upper thresholds of 0.2 and 0.28, corresponding to a probability-ratio interval of $[0.8,1.28]$, and a dual-clipping coefficient of 3.0. No KL penalty is included in either the reward or the optimization loss.

We use AdamW with a constant learning rate of $1\times10^{-5}$, no warmup, weight decay of 0.01, and random seed 42. Training runs for five epochs, with checkpoints saved every 16 updates. We select the checkpoint with the highest answer accuracy on the independent validation set described in Appendix~\ref{app:rl_data}, yielding the checkpoint at update 128 used for all reported ASPECT-RL results.

\begin{algorithm}[t]
\caption{Three-stage supervised fine-tuning}
\label{alg:sft}
\begin{algorithmic}[1]
\Require Backbone model; QA records $\mathcal{D}_{\mathrm{QA}}$; visual-supervision records $\mathcal{D}_{\mathrm{V}}$
\Require Cached teacher targets; stage lengths $(T_1,T_2,T_3)=(1000,1000,1500)$
\State Initialize LoRA adapters, visual-token parameters, and visual-supervision modules
\For{stage $s\in\{1,2,3\}$}
    \If{$s=2$}
        \State $\mathcal{D}_s \gets \mathcal{D}_{\mathrm{V}}\cup\mathcal{D}_{\mathrm{QA}}$
    \Else
        \State $\mathcal{D}_s \gets \mathcal{D}_{\mathrm{QA}}$
    \EndIf
    \For{$t=1,\ldots,T_s$}
        \State Sample a batch from $\mathcal{D}_s$
        \If{$s=1$} \Comment{Perceive}
            \State Input $\gets$ image, question, and visual-token block
            \State Target $\gets$ answer block
        \ElsIf{$s=2$} \Comment{Generate}
            \State Input $\gets$ image and feature query
            \State Target $\gets$ visual-token block
        \Else \Comment{Reason}
            \State Input $\gets$ image and question
            \State Target $\gets$ \texttt{<think>} with visual tokens, \texttt{<observe>}, and \texttt{<answer>}
        \EndIf
        \State Run a teacher-forced forward pass and collect visual-token hidden states
        \State Compute $\mathcal{L}_{\mathrm{LM}}$ over all target tokens, masking input positions
        \State Compute $\mathcal{L}_{\mathrm{uni}}$, $\mathcal{L}_{\mathrm{align}}$, and $\mathcal{L}_{\mathrm{count}}$ from available teacher targets
        \State Set unavailable visual-loss terms to zero
        \State $\mathcal{L}\gets\mathcal{L}_{\mathrm{LM}}+\lambda_{\mathrm{u}}\mathcal{L}_{\mathrm{uni}}+\lambda_{\mathrm{a}}\mathcal{L}_{\mathrm{align}}+\lambda_{\mathrm{c}}\mathcal{L}_{\mathrm{count}}$
        \State Update trainable parameters using AdamW
    \EndFor
\EndFor
\State \Return ASPECT-SFT
\end{algorithmic}
\end{algorithm}

\subsection{Training Algorithms and Inference}
\label{app:algorithms}
We summarize the SFT and RL procedures in Algorithms~\ref{alg:sft} and~\ref{alg:rl}, respectively. The algorithms specify how supervision and generated responses enter training; optimization settings are provided in Table~\ref{tab:training_config} and Appendices~\ref{app:sft_implementation}--\ref{app:rl_implementation}.

\begin{algorithm}[t]
\caption{Reinforcement learning with answer--observation consistency}
\label{alg:rl}
\begin{algorithmic}[1]
\Require ASPECT-SFT policy $\pi_\theta$; RL dataset $\mathcal{D}_{\mathrm{RL}}$; validation set $\mathcal{D}_{\mathrm{val}}$
\Require Reward coefficients $\beta,\gamma$; responses per question $K=8$
\State Initialize the RL policy from ASPECT-SFT
\For{each of five training epochs}
    \For{each batch of 48 questions from $\mathcal{D}_{\mathrm{RL}}$}
        \For{each image--question pair $(x,u)$ with reference answer $a^\star$}
            \State Sample responses $\{y_i\}_{i=1}^{K}$ from $\pi_\theta(\cdot\mid x,u)$
            \For{each response $y_i$}
                \State Check structural validity $V_s(y_i)$
                \If{$V_s(y_i)=0$}
                    \State $R_i \gets -\gamma$
                \Else
                    \State Extract reported counts $C_i$ and final answer $\hat a_i$
                    \State $\mathrm{Acc}_i \gets \mathbf{1}[\hat a_i=a^\star]$
                    \State $\mathrm{Con}_i\gets\mathbf{1}[g_u(C_i)=\pi_u(\hat a_i)]$, or $0$ if undetermined
                    \State $R_i \gets \mathrm{Acc}_i+\beta\,\mathrm{Con}_i$
                \EndIf
            \EndFor
            \State Normalize group rewards to obtain relative advantages
        \EndFor
        \State Update visual and language LoRA parameters with GRPO
        \State Use asymmetric clipping and no KL penalty as specified in Appendix~\ref{app:rl_implementation}
        \State Save a checkpoint every 16 training steps
    \EndFor
\EndFor
\State \Return The saved checkpoint with the highest accuracy on $\mathcal{D}_{\mathrm{val}}$
\end{algorithmic}
\end{algorithm}

\paragraph{Inference.}
ASPECT generates visual tokens, textual observations, and the final answer autoregressively from the image and question. The frozen teachers and visual-supervision modules are not used at inference. Reported counts are generated as part of the response, rather than read from the auxiliary count head, which provides image-wide supervision during SFT. The model determines the generated token sequence without an external routing mechanism.

\section{Evaluation Protocols and Metric Definitions}
\label{app:evaluation}

\subsection{Benchmarks and Model Evaluation Settings}
\label{app:evaluation_settings}

\paragraph{Evaluation datasets.}
We evaluate ASPECT and all baselines on PathoVernier and three external pathology benchmarks. For PathCLS, we use all 1,632 questions from the official PathMMU PathCLS test subset, with the original multiple-choice options and answer labels. PathMMU data are not used for SFT or checkpoint selection. For PathVQA and Quilt-VQA, we evaluate native closed-ended yes/no questions from their official test splits, retaining only H\&E histology images. Each unique image is assessed independently by GPT-5.5 and Gemini-3.1-Pro and retained if either model identifies it as H\&E histology. This filtering retains 1,306 of 3,362 PathVQA questions and 314 of 343 Quilt-VQA questions. All models are evaluated on the same retained subsets. These evaluations use no in-context examples; PathVQA training examples used for SFT are described in Appendix~\ref{app:sft_data}.

\begin{table*}[h!]
\renewcommand{\arraystretch}{1.1}
\setlength{\tabcolsep}{6pt}
\centering
\caption{Evaluation subsets and output token limits for local models and closed-source APIs.}
\resizebox{\textwidth}{!}{
\begin{tabular}{llrcc}
\toprule[1pt]
Benchmark & Evaluation subset & Questions & Local token limit & API token limit \\
\midrule[0.7pt]
PathoVernier & Full benchmark & 759 & 512 & 512 \\
PathCLS & Official test & 1,632 & 512 & 512 \\
PathVQA & H\&E, closed-ended test & 1,306 & 512 & 512 \\
Quilt-VQA & H\&E, closed-ended test & 314 & 512 & 512 \\
\bottomrule[1pt]
\end{tabular}
}
\label{tab:evaluation_settings}
\end{table*}

\paragraph{Inference settings.}
Images are resized to $512\times512$ pixels before processing with each model's native image processor. We generate one response per question. Local models use greedy decoding with sampling disabled and seed 42; closed-source APIs use temperature zero. Output token limits are listed in Table~\ref{tab:evaluation_settings}. The closed-source model identifiers are \texttt{gpt-5.5} and \texttt{gemini-3.1-pro-preview}. For PathoVernier and PathCLS, GPT-5.5 was evaluated on September 2, 2026, and Gemini-3.1-Pro on September 3, 2026. Both models were evaluated on PathVQA and Quilt-VQA on September 15, 2026.

\paragraph{Prompts and scoring.}
On PathoVernier, all models receive identical questions and answer options. Baselines additionally receive a system prompt requesting whole-image cell-type counts, the quantities required by the question, and a final answer, whereas ASPECT uses its trained response format without a system prompt. All responses are scored with the same parser. PathVQA and Quilt-VQA request only yes/no answers and use no system prompt. PathCLS requests a brief morphological justification followed by the selected option letter. Intermediate counts are not required for these external tasks. Unparseable final answers are scored as incorrect. Failed API requests are retried up to four times, and unresolved failures are also scored as incorrect. Metric definitions and eligibility criteria are detailed in Appendix~\ref{app:metric_protocols}, and the evaluation prompts are provided in Appendix~\ref{app:evaluation_prompts}.

\subsection{Task Rules, Metrics, and Eligible Responses}
\label{app:metric_protocols}

\paragraph{Task rules.}
Reference counts are obtained by assigning annotated nuclei to the geometric regions specified in each question according to their centroids. Region selection identifies the region with the largest count of the queried cell type among equal-area regions. Region comparison compares one cell type across two equal-area regions, while cell-type comparison compares two types within the image. Let $a$ and $b$ denote the first and second counts in the comparison. For $b>0$, both tasks assign the ratio $r=a/b$ to a category using
\begin{equation}
h(r)=
\begin{cases}
\texttt{much\_fewer}, & r\leq0.5,\\
\texttt{comparable}, & 0.5<r<2.0,\\
\texttt{much\_more}, & r\geq2.0.
\end{cases}
\label{eq:app_comparison_rule}
\end{equation}
When $b=0$, the result is \texttt{much\_more} if $a>0$ and \texttt{comparable} if $a=0$. These cases remain determinable. For equal-area regions, the count ratio equals the density ratio.

Multi-step composition first selects the region with the largest count of the queried type and then assigns that type's proportion $p$ within the selected region to a category:
\begin{equation}
h_{\mathrm{prop}}(p;t_1,t_2)=
\begin{cases}
\texttt{low}, & p<t_1,\\
\texttt{mid}, & t_1\leq p<t_2,\\
\texttt{high}, & p\geq t_2.
\end{cases}
\label{eq:app_proportion_rule}
\end{equation}
The thresholds $t_1$ and $t_2$ are estimated from training-split tertiles separately for each cell type and spatial partition, rounded to increments of $0.05$, and then frozen. Each question states its applicable thresholds explicitly. Table~\ref{tab:proportion_thresholds} lists the thresholds for quadrants, four horizontal bands, and four vertical bands. Tumor is excluded from this task because its median proportion is approximately one. Candidate questions whose reference proportions fall within a $5\%$ relative margin of either threshold are excluded during construction.

\begin{table*}[h!]
\renewcommand{\arraystretch}{1.1}
\setlength{\tabcolsep}{6pt}
\centering
\caption{Frozen proportion thresholds $(t_1,t_2)$ for multi-step composition, estimated separately for each spatial partition and cell type.}
\resizebox{0.7\textwidth}{!}{
\begin{tabular}{lcccc}
\toprule[1pt]
Partition & Stromal-like & Lymphocyte & Epithelial-like & Inflammatory \\
\midrule[0.7pt]
Quadrants & $(0.50,0.85)$ & $(0.35,0.55)$ & $(0.60,0.90)$ & $(0.55,0.80)$ \\
Horizontal bands & $(0.40,0.80)$ & $(0.30,0.50)$ & $(0.55,0.80)$ & $(0.55,0.75)$ \\
Vertical bands & $(0.40,0.75)$ & $(0.30,0.50)$ & $(0.60,0.85)$ & $(0.55,0.75)$ \\
\bottomrule[1pt]
\end{tabular}
}
\label{tab:proportion_thresholds}
\end{table*}

If reported counts yield a tied maximum in region selection or the first step of multi-step composition, the count-derived region is undeterminable. Such responses are excluded from CA and Consistency, but remain eligible for Acc and Count Acc; RAWR eligibility follows its correct-answer and complete-count requirements. A tied maximum therefore differs from a zero denominator in a comparison task, for which the rule above returns a definite category.

\paragraph{Answer and count accuracy.}
Let $\hat a$ and $a^\star$ denote the predicted and reference answers, respectively. Answer accuracy evaluates the complete answer, including both the region and proportion category for multi-step composition. An unparseable final answer is scored as incorrect. To evaluate the reported measurements, let $K$ index the counts required by a question, $q_i$ denote reference count $i$, and $\operatorname{dom}(C)$ denote the reported count keys. The per-question count score is
\begin{equation}
S(C,Q)
=
\frac{1}{|K|}
\sum_{i\in K\cap\operatorname{dom}(C)}
\mathbf{1}\!\left[
|C_i-q_i|\leq \max(1,0.1q_i)
\right],
\label{eq:app_count_score}
\end{equation}
where $Q=\{q_i\}_{i\in K}$ is the set of required reference counts and $\mathbf{1}[\cdot]$ is the indicator function. Each required count contributes equally, and missing counts contribute zero. We report
\begin{equation}
\mathrm{Acc}
=
\mathbb{E}\!\left[\mathbf{1}[\hat a=a^\star]\right],
\qquad
\mathrm{Count\ Acc}
=
\mathbb{E}\!\left[S(C,Q)\right].
\label{eq:app_accuracy}
\end{equation}
Both averages include all evaluation questions, with equal weight per question. Count Acc therefore measures the fraction of required counts within tolerance rather than exact integer-count accuracy.

\paragraph{Gold-referenced coherent accuracy.}
CA evaluates whether the reported counts imply the reference task conclusion. Let $g_u(C)$ denote the conclusion checked by CA and $\pi_u(a^\star)$ the corresponding component of the reference answer:
\begin{equation}
\mathrm{CA}
=
\mathbb{E}\!\left[
\mathbf{1}[g_u(C)=\pi_u(a^\star)]
\,\middle|\,
g_u(C)\text{ is defined}
\right].
\label{eq:app_ca}
\end{equation}
For region selection and the two comparison tasks, this check uses the complete task conclusion. For multi-step composition, $g_u$ checks only region selection, and $\pi_u$ extracts the reference region. Its eligibility therefore depends on the counts needed to select a region, without requiring the additional counts needed to determine the proportion category. CA does not require the model's final answer to be correct. Moreover, counts can imply the correct conclusion while remaining numerically inaccurate, so CA and Count Acc measure different properties.

\paragraph{Right answer, wrong reason.}
RAWR evaluates counting errors among responses that have a correct final answer and report every required count:
\begin{equation}
\mathrm{RAWR}
=
1-
\mathbb{E}\!\left[
S(C,Q)
\,\middle|\,
\hat a=a^\star,\;
K\subseteq\operatorname{dom}(C)
\right].
\label{eq:app_rawr}
\end{equation}
Thus, Count Acc averages count agreement over all questions, whereas RAWR averages the fraction of counts outside tolerance over correct responses with complete measurements. RAWR is not a binary indicator of whether a response contains any counting error.

\paragraph{Consistency with the model's own answer.}
Consistency evaluates whether the conclusion implied by the reported counts agrees with the corresponding component of the model's own answer. Using the same task check $g_u$ and answer projection $\pi_u$ as CA, we define
\begin{equation}
\mathrm{Consistency}
=
\mathbb{E}\!\left[
\mathbf{1}[g_u(C)=\pi_u(\hat a)]
\,\middle|\,
g_u(C)\text{ and }\pi_u(\hat a)\text{ are defined}
\right].
\label{eq:app_consistency}
\end{equation}
CA compares the count-derived conclusion with $\pi_u(a^\star)$, whereas Consistency compares it with $\pi_u(\hat a)$. For multi-step composition, both checks evaluate region selection only and do not require the proportion denominator. The full region-and-proportion answer is evaluated by Acc, while $S(C,Q)$ measures agreement between reported and reference counts. Consequently, a response can be consistent despite an incorrect proportion category or an incorrect final answer. The indicator in ~\eqref{eq:app_consistency} is the same as $\mathrm{Con}$ in the RL reward: undeterminable cases receive zero consistency reward during training but are excluded from the evaluation metric's denominator. The off-diagonal categories in Figure~5 therefore do not measure the rate of inconsistency with the model's own answer.

\paragraph{Eligible responses.}
Acc and Count Acc include all evaluation questions. CA includes responses with a determinable count-derived conclusion even when the final answer is missing or unparseable. RAWR requires a correct final answer and complete reported counts, whereas Consistency requires both the count-derived conclusion and the corresponding predicted-answer component to be determinable. Counts are scored field by field using the evaluation parser's output; negative values or duplicate keys do not automatically assign zero to the entire response. The training validity gate $V_s$ is not applied as a response-level filter during evaluation. Missing required counts contribute zero to Count Acc. Count coverage is the fraction of all 759 questions for which the reported counts yield a determinable conclusion under the task rules used for CA. For multi-step composition, this requires determining the selected region, irrespective of the proportion band. Complete-count coverage instead requires every necessary count to be reported, regardless of answer correctness. Complete counts can still yield an undeterminable conclusion when regional maxima are tied. In Table \ref{table1}, CA and RAWR are omitted when CA coverage is below 30\%; this reporting threshold does not change either metric's definition or eligible sample set. Table~\ref{tab:metric_coverage} reports both coverage measures, RAWR sample counts, and unconditional Count Acc.

\begin{table*}[h!]
\renewcommand{\arraystretch}{1.1}
\setlength{\tabcolsep}{6pt}
\centering
\caption{Evaluation coverage and unconditional count accuracy on PathoVernier. Both coverage measures report counts out of all 759 questions. $N_{\mathrm{RAWR}}$ is the number of correct responses with complete reported counts. Count Acc averages $S(C,Q)$ over all questions, including failures and incomplete responses.}
\resizebox{0.8\textwidth}{!}{
\begin{tabular}{lcccc}
\toprule[1pt]
Model & CA coverage & Complete-count coverage & $N_{\mathrm{RAWR}}$ & Count Acc$\uparrow$ \\
\midrule[0.7pt]
GPT-5.5                & 748/759 & 749/759 & 447 & 0.210 \\
Gemini-3.1-Pro         & 733/759 & 750/759 & 472 & 0.296 \\
Qwen3-VL-8B            & 233/759 & 261/759 & 106 & 0.068 \\
Qwen3-VL-32B           & 682/759 & 700/759 & 333 & 0.178 \\
InternVL3-8B           & 448/759 & 449/759 & 166 & 0.071 \\
InternVL3-38B          & 750/759 & 753/759 & 340 & 0.153 \\
Quilt-LLaVA-7B         & 1/759   & 2/759   & 1   & 0.000 \\
PathGen-LLaVA-13B      & 0/759   & 0/759   & 0   & 0.000 \\
HuatuoGPT-Vision-7B    & 12/759  & 12/759  & 2   & 0.002 \\
Patho-R1-7B            & 0/759   & 0/759   & 0   & 0.000 \\
\midrule[0.7pt]
ASPECT-8B (SFT)        & 714/759 & 738/759 & 525 & 0.444 \\
ASPECT-8B (RL)         & 748/759 & 752/759 & 563 & \textbf{0.471} \\
\bottomrule[1pt]
\end{tabular}
}
\label{tab:metric_coverage}
\end{table*}

ASPECT achieves the highest Count Acc over all questions (0.471 versus 0.296 for Gemini-3.1-Pro), showing that its improvement in measurement accuracy extends beyond the conditional subsets used for CA and RAWR.

\subsection{Ablation and Image-Mismatch Protocols}
\label{app:ablation_protocols}

\paragraph{Visual supervision and training curriculum.}
We evaluate the SFT ablations before RL to separate the effects of supervised training from reward optimization. Starting from the same backbone, we remove pathology feature reconstruction, cell feature alignment, or count supervision individually, and additionally evaluate a variant without all three visual losses. These loss-removal variants retain the visual tokens, response formats, three-stage curriculum, and remaining loss weights. Assistant-token cross-entropy has weight one throughout training. Direct Reason retains all three visual losses and trains on the 13,095 QA examples using the Reason-stage response format from the first update. It runs for 3,500 updates, matching the total update budget of the complete curriculum. Unlike the full curriculum, whose Generate stage also includes visual-supervision examples, Direct Reason uses QA examples throughout. All variants follow the evaluation protocol in Appendix~\ref{app:evaluation_settings}.

\paragraph{Reinforcement learning rewards.}
The reward ablations initialize from the same ASPECT-SFT checkpoint and use the same 1,372-question RL set and GRPO configuration. We retain the structural validity gate and invalid-response penalty in every variant:
\begin{equation}
R=V_s\,r-0.1(1-V_s),
\label{eq:app_reward_ablation}
\end{equation}
where the valid-response reward $r$ is $\mathrm{Acc}$, $0.5\,\mathrm{Con}$, $\mathrm{Acc}+0.5\,S(C,Q)$, or $\mathrm{Acc}+0.5\,\mathrm{Con}$ for answer-only, consistency-only, answer-plus-count, and ASPECT RL, respectively. The count-reward variant uses the per-question tolerance score defined in ~\eqref{eq:app_count_score}, rather than the dataset-level Count Acc. The SFT row provides the initialization baseline without RL. Checkpoints are selected by answer accuracy on the independent validation set.

\paragraph{Image mismatch.}
Figure~\ref{fig5}(a) evaluates the model without auxiliary visual supervision, ASPECT-SFT, and ASPECT-RL using original and mismatched images. We construct a permutation of unique patches within each source dataset, requiring every patch to map to a different patch. This preserves dataset provenance while breaking the correspondence between the image and question. All questions associated with an original patch share the same replacement image, and all three models use identical mappings. We generate three mappings with seeds 0, 1, and 2 over the initial 786-question candidate set and evaluate only the final 759 benchmark questions. Questions, options, reference answers, and reference counts remain unchanged. We report the mean mismatched-image Acc and Count Acc across the three mappings; all corresponding standard deviations are at most $0.017$. This variation reflects image replacements rather than training runs.

\paragraph{Joint outcome analysis.}
Figure~\ref{fig5}(b) decomposes the responses of ASPECT-SFT, answer-only RL, and ASPECT-RL according to final-answer correctness and the gold-referenced count check. For each response, we define
\begin{equation}
A=\mathbf{1}[\hat a=a^\star],
\qquad
B=\mathbf{1}[g_u(C)=\pi_u(a^\star)],
\label{eq:app_joint_outcomes}
\end{equation}
where $g_u$ and $\pi_u$ follow Appendix~\ref{app:metric_protocols}. Responses with a determinable count-derived conclusion are assigned to the four $(B,A)$ combinations; responses for which $g_u(C)$ is undeterminable form a fifth category. All category proportions use the full 759-question benchmark as their denominator. For multi-step composition, $B$ evaluates region selection, whereas $A$ evaluates the complete region-and-proportion answer. Thus, $B=1,A=0$ includes responses that select the correct region but give an incorrect proportion category. This group need not be inconsistent with the model's own reported counts, and the off-diagonal proportion is not the complement of Consistency.

\section{Prompts and Response Formats}
\label{app:prompts}

\subsection{Training Data Generation}
\label{app:data_generation_prompts}

We generate the textual content of SFT question-answer records using GPT-5.6-sol. Each request includes the H\&E image, question, reference answer, and source-specific context. The five nucleus-annotated datasets provide reference cell counts, whereas cellular NCT-CRC-HE-100K patches provide pseudo-counts together with their tissue labels. PathVQA supplies the original yes/no answer without a nuclear composition input. All sources share the system prompt below. The displayed prompts are excerpts, and braces denote sample-specific fields.

\begin{promptbox}{GPT-5.6-sol: Shared system prompt}
You are an expert pathologist. You are given an H\&E image, the GROUND-TRUTH nuclear composition of the field, a question, and the CORRECT answer. Your job is NOT to decide the answer (it is already given and correct).

Produce EXACTLY two lines:

\texttt{OBSERVE:} one sentence describing the overall visual appearance of the field --- cellularity, tissue architecture, dominant nuclear morphology --- WITHOUT stating any numeric counts (no digits).

\texttt{REASON:} 2-3 sentences that justify the GIVEN answer, citing counts ONLY from the provided composition, leading to that answer.

Rules: never invent numbers; OBSERVE must contain NO digits; do NOT repeat the OBSERVE sentence inside REASON \ldots; do NOT write `FINAL:' or the answer token --- that is appended automatically.
\end{promptbox}

The source-specific user inputs are summarized below. For the five nucleus-annotated datasets, the composition includes explicitly supplied zero counts for absent types. CRC inputs distinguish cellular patches, which supply detected composition, from non-cellular patches. PathVQA instead requests an explanation based on image morphology.

\begin{promptbox}{GPT-5.6-sol: Source-specific user inputs}
\textbf{Five nucleus-annotated datasets}

Tissue/organ: \texttt{\{organ\}} (H\&E).

Ground-truth nuclear composition of THIS field: \texttt{\{composition\}}.

Question: \texttt{\{question\}}

CORRECT answer: \texttt{\{gold\}}

Output the two lines now:

\texttt{OBSERVE: ...}

\texttt{REASON: ...}

\tcblower

\textbf{NCT-CRC-HE-100K}

H\&E colorectal histology.

Ground-truth tissue class: \texttt{\{option\}} (\texttt{\{tissue\_class\}}).

\emph{For cellular patches:}

Detected nuclear composition: \texttt{\{composition\}}.

\emph{For non-cellular patches:}

This field is essentially acellular (no significant nuclei).

Question: \texttt{\{question\}}

CORRECT answer: \texttt{\{gold\}}

Output the two lines now:

\texttt{OBSERVE: ...}

\texttt{REASON: ...}

\medskip
\hrule
\medskip

\textbf{PathVQA}

H\&E pathology microscopy image.

Question: \texttt{\{question\}}

CORRECT answer: \texttt{\{yes/no\}}

Write two lines. OBSERVE = one sentence on the overall appearance relevant to the question (no digits). REASON = 1-3 sentences that justify the given yes/no answer from the morphology.

\texttt{OBSERVE: ...}

\texttt{REASON: ...}
\end{promptbox}

The generated lines are incorporated into the \texttt{<observe>} and \texttt{<answer>} blocks. Where count observations are included, their JSON is inserted programmatically from the corresponding annotations or pseudo-labels. PathVQA observations contain qualitative descriptions without count JSON. The \texttt{FINAL:} line is appended from the reference answer.

For counting questions, GPT-5.6-sol subsequently rewrites the reasoning notes in \texttt{<answer>}, including their count statements, conclusions, and final-answer lines. It receives text only, with multiple notes grouped by identifiers. Rewrites must preserve the numerical multiset and the final-answer line; failures revert to the template version.

\begin{promptbox}{GPT-5.6-sol: Answer-note rewriting}
You are rewriting the wording of short analytical notes that accompany histopathology counting questions. Each note states some counts and draws a conclusion. Rewrite EACH note below.

Hard requirements:
\begin{itemize}
    \item Keep every number EXACTLY as given. Do not add, drop, round or recompute any number. You cannot see the image, so you have no basis to change any count.
    \item Keep the final line exactly as \texttt{FINAL: \textless option\textgreater} with the option unchanged.
    \item Vary sentence structure, connectives and register \ldots; do not reuse one skeleton with different numbers \ldots
    \item Keep it to one or two sentences plus the FINAL line.
    \item Output \ldots one rewritten note per input, each introduced by \texttt{\#\#\# \textless id\textgreater} \ldots
\end{itemize}

Notes to rewrite:

\texttt{\#\#\# \{id\}}

\texttt{\{answer\_note\_including\_FINAL\}}

\ldots
\end{promptbox}

\subsection{Benchmark Question Rewriting}
\label{app:question_rewriting}

We use GPT-5.6-sol to diversify question wording while preserving the queried quantities. Training and benchmark questions use separate phrasing pools, denoted A and B, with no template shared between them. The generator receives a reference template, its placeholder list, and an explicit description of the intended measurement. Numerical rules, reference counts, and final answers are handled programmatically rather than generated by the paraphrasing model.

\begin{promptbox}{GPT-5.6-sol: Question paraphrasing}
You are helping diversify the phrasing of questions in a computational-pathology benchmark.

Reference question template (Python format string): \texttt{\{ref\}}

Placeholders: \texttt{\{ph\}}

What the question asks: \texttt{\{meaning\}}

Write \texttt{\{n\}} ALTERNATIVE phrasings of the SAME question. Hard requirements:
\begin{itemize}
    \item Use EXACTLY the same placeholders, each exactly once. Do not add or drop any.
    \item Ask for exactly the same quantity. Do not change what is being compared, which region, or how it is normalised.
    \item Do NOT hint at the answer: the question must not contain comparative or magnitude words such as more/fewer/higher/lower/most/least/high/low/dense.
    \item Vary sentence structure and register (some clinical, some plain), not just synonyms.
    \item One per line, no numbering, no quotes, nothing else.
\end{itemize}
\end{promptbox}

Here, \texttt{\{ref\}} is the reference template, \texttt{\{ph\}} lists its placeholders, \texttt{\{meaning\}} specifies the quantity being queried, and \texttt{\{n\}} is the requested number of alternatives. Within question templates, \texttt{\{t\}} denotes the queried nucleus type, \texttt{\{t1\}} and \texttt{\{t2\}} denote two compared types, and \texttt{\{r1\}} and \texttt{\{r2\}} denote two compared regions. The placeholder \texttt{\{region\}} denotes the whole-image field. The following example specifies that both steps of a multi-step composition question concern the same nucleus type.

\begin{promptbox}{Example generator input: Multi-step composition}
\textbf{Reference template}

Determine the region whose \texttt{\{t\}} nuclear density is greatest, then state the proportion of that region's nuclei that are \texttt{\{t\}}.

\textbf{Placeholder list}

\texttt{\{t\}}

\textbf{Intended measurement}

first which equal-area region contains the greatest density of nucleus type \texttt{\{t\}}, then what proportion of ALL nuclei in THAT region are of the SAME type \texttt{\{t\}}. Only one nucleus type is involved throughout --- do not introduce a second type.
\end{promptbox}

Candidates first undergo deterministic checks for placeholder identity and prohibited directional wording. Geometric-region questions must retain the word ``region'', use ``nuclei'' rather than ``cells'', and restrict ``area'' to the expression ``per unit area''. Cross-pool Jaccard similarity must not exceed $0.55$. Candidates passing these checks are evaluated by GPT-5.4, using the reference and candidate templates instantiated with the same placeholder values.

\begin{promptbox}{GPT-5.4: Independent semantic verification}
Two questions about the same histopathology image are shown.

A: \texttt{\{a\}}

B: \texttt{\{b\}}

Do A and B ask for EXACTLY the same quantity --- same object type, same region, same normalisation (per-area vs raw count), same comparison direction? A difference in wording is fine; a difference in what is measured is not.

Answer with ONLY one word: SAME or DIFFERENT.
\end{promptbox}

The candidate is accepted when the verifier returns \texttt{SAME} and rejected when it returns \texttt{DIFFERENT}. Failed candidates are regenerated, with the original template retained when retries are exhausted. Rewriting changes question wording without modifying the numerical keys in \texttt{\textless observe\textgreater} or the programmatically determined \texttt{FINAL} answer. Independent answer verification and expert review follow the procedures in Appendix~\ref{app:benchmark_construction}.

\subsection{SFT and RL Prompts}
\label{app:training_prompts}

We present the user inputs and assistant targets used in the three SFT stages, followed by the prompt used for RL. Images precede text in every input. For compactness, \texttt{[UNI block]} denotes \texttt{\textless|anchor\_start|\textgreater}, eight consecutive \texttt{\textless|uni\_pad|\textgreater} tokens, and \texttt{\textless|anchor\_end|\textgreater}; \texttt{[CELL block]} uses the same delimiters with six consecutive \texttt{\textless|cell\_pad|\textgreater} tokens. These bracketed names are display abbreviations and are replaced by the corresponding token sequences in training. The UNI block always precedes the CELL block. SFT applies cross-entropy to all assistant-target tokens with unit weight and masks user-prompt positions.

\begin{promptbox}{Perceive: Answer from visual tokens}
\textbf{User input}\par
\texttt{\textless image\textgreater}the patch feature of the image is \texttt{[UNI block]}, and the cell composition of the image is \texttt{[CELL block]}. \texttt{\{QUESTION\}}

\medskip
\textbf{Assistant target}\par
\texttt{\{ANSWER\}}
\end{promptbox}

Here, \texttt{\{ANSWER\}} is the prepared answer block, including its \texttt{\textless answer\textgreater} and \texttt{\textless/answer\textgreater} delimiters. The target contains neither a \texttt{\textless think\textgreater} block nor an \texttt{\textless observe\textgreater} block. Visual tokens occur in the user input and therefore receive visual supervision but no token-prediction loss at their input positions.

\begin{promptbox}{Generate: Predict visual tokens}
\textbf{User input}\par
\texttt{\textless image\textgreater}What is the patch feature, cell composition of the image?

\medskip
\textbf{Assistant target}\par
\texttt{[UNI block][CELL block]}
\end{promptbox}

Generate uses both visual-supervision records and QA records, replacing their questions with the feature query above. Every record supervises prediction of the visual-token sequence through cross-entropy, together with the available visual supervision losses.

\begin{promptbox}{Reason: Generate observations and an answer}
\textbf{User input}\par
\texttt{\textless image\textgreater\{QUESTION\}}

\medskip
\textbf{Assistant target}\par
\texttt{\textless think\textgreater} the patch feature of the image is \texttt{[UNI block]}, and the cell composition of the image is \texttt{[CELL block]}. \texttt{\textless/think\textgreater}\par
\texttt{\textless observe\textgreater\{OBSERVATION\}\textless/observe\textgreater}\par
\texttt{\textless answer\textgreater} \texttt{\{REASONING\}}\par
\texttt{FINAL: \{OPTION\} \textless/answer\textgreater}
\end{promptbox}

The Reason target places the visual tokens inside the \texttt{\textless think\textgreater} block before the observation and answer. The \texttt{\{OBSERVATION\}} field contains the descriptive sentence and, where provided by the data source, the count JSON described in Appendix~\ref{app:data_generation_prompts}. PathVQA observations contain only descriptive prose. The question includes its answer options.

\begin{promptbox}{Reinforcement learning: Sample a response}
\textbf{System prompt}\par
None.

\medskip
\textbf{User input}\par
\texttt{\textless image\textgreater}\par
\texttt{\{QUESTION\}}\par
Choose exactly one of: \texttt{\{opt1, opt2, \ldots\}}

\medskip
\textbf{Assistant response}\par
Sampled from the policy; no fixed target is supplied.
\end{promptbox}

RL retains the image-and-question input structure of Reason and explicitly lists the available options. The policy generates its response in the format learned during SFT, without additional formatting instructions in the prompt. Responses are scored using the reward described in Appendix~\ref{app:rl_implementation}.

For training samples marked as non-cellular, the CELL block and the corresponding cell-composition clause are omitted. Their Generate query is ``What is the patch feature of the image?'', and its target contains only the UNI block. This adjustment applies to training examples; inference uses no external routing to select which visual-token blocks the model generates.

\subsection{Evaluation Prompts}
\label{app:evaluation_prompts}

We provide the evaluation prompts below. Each question is accompanied by its image; image inputs are omitted from the displayed templates. Placeholders are replaced with the question and its available answer options. All evaluations use zero-shot prompting without demonstration examples.

\paragraph{PathoVernier.}
Baseline models receive explicit instructions to report nuclear counts before their final answer. ASPECT uses no system prompt and follows the response format learned during SFT, with the image, question, and available options as input, as illustrated in Appendix~\ref{app:training_prompts}. All models receive the same questions and answer options.

\begin{promptbox}{PathoVernier: Baseline evaluation}
\textbf{System prompt}\par
You are an expert pathologist analyzing an H\&E-stained histology image. You MUST base your answer strictly on what is visible in THIS image: actually look at and count the nuclei you can see. Never answer from priors, definitions, or the wording of the question alone --- if you did not look at the image, you cannot answer. Think step by step. Then output ONE JSON object wrapped in \texttt{\textless observe\textgreater...\textless/observe\textgreater}: FIRST the whole-image count of each main nucleus type you can see --- epithelial-like, stromal-like, lymphocyte, inflammatory, and tumor separately when identifiable (do NOT report a type that is not present in this patch) --- THEN the specific quantities the question asks you to report. Finally, output a line `FINAL: X' where X is EXACTLY one of the allowed option tokens, copied verbatim.

\medskip
\textbf{User prompt}\par
\texttt{\{question\}}, Choose exactly one of these option tokens: \texttt{\{opt1, opt2, \ldots\}}. Respond with your reasoning, then \texttt{\textless observe\textgreater\{...\}\textless/observe\textgreater}, then a final line `\texttt{FINAL: \textless token\textgreater}'.
\end{promptbox}

\paragraph{PathCLS.}
The prompt asks the model to identify the tissue or lesion, briefly justify its choice from the image, and end with the selected option letter. It does not require structured count observations.

\begin{promptbox}{PathCLS: Tissue and lesion classification}
\textbf{System prompt}\par
You are an expert pathologist analyzing a single H\&E-stained histology image. This is a MULTIPLE-CHOICE QUESTION about which tissue or lesion type the image shows. You may first note what you observe about the cells, but your final \texttt{\textless answer\textgreater} MUST directly address THIS question: decide which of the listed options best matches the tissue/lesion in the image, briefly justify it from the morphology, and end with a line `FINAL: X' where X is EXACTLY one option letter. Do NOT just describe cell counts --- you must pick the correct option.

\medskip
\textbf{User prompt}\par
\texttt{\{question\}}\par
Options: A) \ldots\ B) \ldots\par
In your answer, choose the single option that correctly names the tissue/lesion shown, and end with `\texttt{FINAL: \textless letter\textgreater}'.
\end{promptbox}

\paragraph{PathVQA and Quilt-VQA.}
Both benchmarks use the same user prompt without a system prompt. Models are asked to return only the binary answer, without reasoning or count observations.

\begin{promptbox}{PathVQA and Quilt-VQA: Closed-ended questions}
\textbf{System prompt}\par
None.

\medskip
\textbf{User prompt}\par
You are looking at a histopathology image.\par
\texttt{\{question\}}\par
Answer with exactly one word: yes or no.
\end{promptbox}

\section{Additional Quantitative Results}
\label{app:additional_results}

\subsection{Results by Task Type}
\label{app:task_results}

\begin{figure*}[h]
\centering
\includegraphics[width=\textwidth]{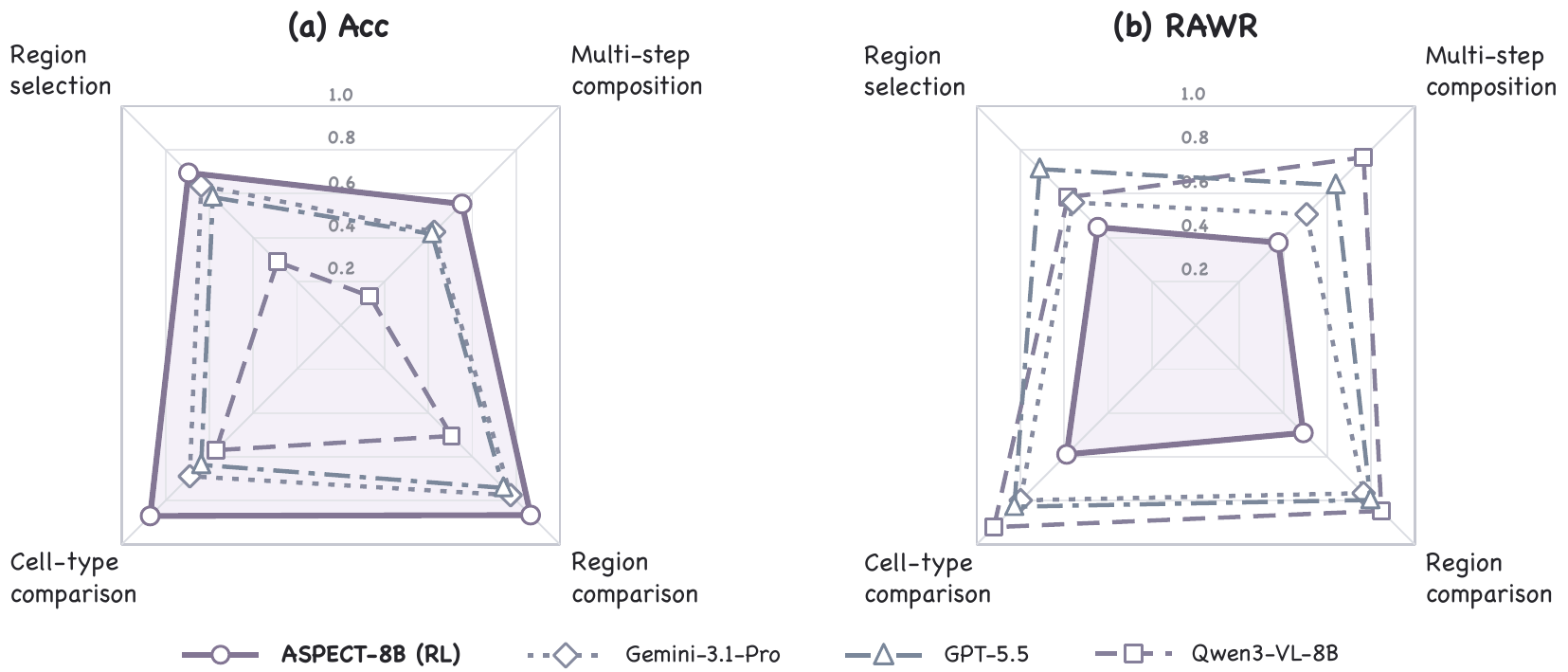}
\caption{Results by task type on PathoVernier. ASPECT achieves higher Acc and lower RAWR than the three comparison models across all four tasks. Larger values are better for Acc; smaller values are better for RAWR.}
\label{fig:task_results}
\end{figure*}

Figure~\ref{fig:task_results} compares ASPECT with its base model and two closed-source models across the four PathoVernier task types. ASPECT achieves higher Acc and lower RAWR in every task, extending its advantage from region selection to multi-step composition and quantitative comparisons. Relative to Gemini-3.1-Pro, its Acc improves by 31.3\% on multi-step composition and 26.3\% on cell-type comparison. Lower RAWR across all four tasks further shows that, among correct answers with complete count observations, ASPECT reports more accurate quantitative evidence. These results connect its gains in task accuracy with improved measurement of the underlying cell composition.

\subsection{Results by Data Source and Imaging Scale}
\label{app:source_results}

\begin{figure*}[t]
\centering
\includegraphics[width=\textwidth]{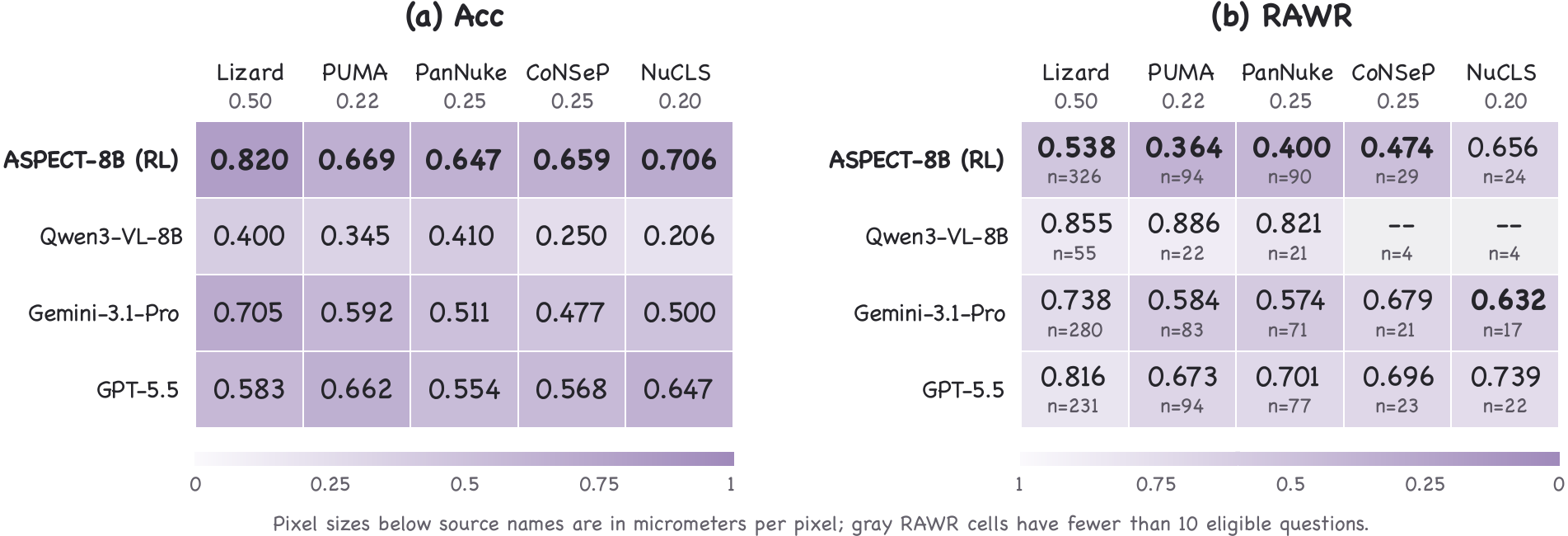}
\caption{Results by source dataset on PathoVernier. Column labels include native pixel sizes in $\mu$m/pixel. Darker shading indicates better performance within each panel. In the RAWR panel, $n$ denotes the number of correct answers with complete count observations; estimates with $n<10$ are withheld. Source and imaging scale are evaluated jointly.}
\label{fig:source_results}
\end{figure*}

Figure~\ref{fig:source_results} breaks down performance across the five source datasets and their native pixel sizes. ASPECT achieves the highest Acc on every source, showing that its gains extend beyond Lizard, which contributes more than half of PathoVernier. On PUMA, ASPECT and GPT-5.5 obtain similar Acc (0.669 versus 0.662), yet ASPECT achieves substantially lower RAWR (0.364 versus 0.673). This comparison illustrates how similar answer accuracy can conceal differences in the quality of reported counts. ASPECT achieves the lowest RAWR on four sources; NuCLS is the exception, where Gemini-3.1-Pro has lower RAWR (0.632 versus 0.656) despite lower Acc. The source-level results thus reveal both the broad improvement in task accuracy and the variation in the accuracy of its supporting measurements.

\subsection{Statistical Uncertainty}
\label{app:statistical_uncertainty}

We quantify test-set sampling uncertainty using 10,000 paired bootstrap resamples of the 553 PathoVernier patches, stratified by source dataset, with random seed 42. Within each source, we sample its original number of patches with replacement and retain all associated questions, including sampling multiplicities. All models share the same resamples. We recompute each metric and its denominator according to Appendix~\ref{app:metric_protocols}, retaining API failures as incorrect answers for Acc and using each model's own eligible samples for CA and RAWR. We report the 2.5th and 97.5th percentiles as 95\% confidence intervals.

As shown in Table~\ref{tab:statistical_uncertainty}, all six paired improvement intervals lie above zero. Against Gemini-3.1-Pro, ASPECT improves Acc by 0.120, with a 95\% CI of [0.077, 0.162]. The positive intervals for CA gains and RAWR reductions further support improvements in both count-derived conclusions and the reported quantities behind correct answers. These comparisons use the paired difference distributions, rather than overlap between individual model intervals.

\begin{table*}[h!]
\renewcommand{\arraystretch}{1.1}
\setlength{\tabcolsep}{6pt}
\centering
\caption{Model scores and paired improvements on PathoVernier with 95\% bootstrap confidence intervals. Improvements are ASPECT minus baseline for Acc and CA, and baseline minus ASPECT for RAWR; positive values favor ASPECT.}
\resizebox{\textwidth}{!}{
\begin{tabular}{lccc}
\toprule[1pt]
Model & Acc$\uparrow$ & CA$\uparrow$ & RAWR$\downarrow$ \\
\midrule[0.7pt]
Qwen3-VL-8B
& 0.374 [0.341, 0.408]
& 0.532 [0.471, 0.595]
& 0.854 [0.801, 0.901] \\
Gemini-3.1-Pro
& 0.626 [0.588, 0.663]
& 0.715 [0.681, 0.749]
& 0.680 [0.651, 0.708] \\
GPT-5.5
& 0.594 [0.558, 0.631]
& 0.652 [0.616, 0.689]
& 0.756 [0.728, 0.784] \\
ASPECT-8B
& \textbf{0.746} [0.714, 0.776]
& \textbf{0.807} [0.777, 0.836]
& \textbf{0.489} [0.460, 0.518] \\
\midrule[1pt]
Comparison & Acc gain & CA gain & RAWR reduction \\
\midrule[0.7pt]
ASPECT vs Qwen3-VL-8B
& +0.372 [+0.327, +0.415]
& +0.275 [+0.207, +0.342]
& +0.365 [+0.305, +0.420] \\
ASPECT vs Gemini-3.1-Pro
& +0.120 [+0.077, +0.162]
& +0.092 [+0.052, +0.132]
& +0.191 [+0.152, +0.229] \\
\bottomrule[1pt]
\end{tabular}
}
\label{tab:statistical_uncertainty}
\end{table*}

\subsection{Sensitivity to Count Tolerance}
\label{app:tolerance_results}

We vary the relative count tolerance $r$ over $\{0, 0.05, \ldots, 0.30\}$ while retaining the absolute tolerance floor of one nucleus. A reported count $c$ agrees with its reference count $q$ when
\begin{equation}
\left|c-q\right| \leq \max\{1,rq\}.
\label{eq:count_tolerance_sensitivity}
\end{equation}
We recompute the count-agreement score and RAWR at each tolerance, keeping each model's correct-answer subset with complete count observations fixed. The main results correspond to $r=0.10$, rather than an average over this tolerance grid. At $r=0$, the criterion still permits an absolute counting error of one.

\begin{figure*}[t]
\centering
\includegraphics[width=\textwidth]{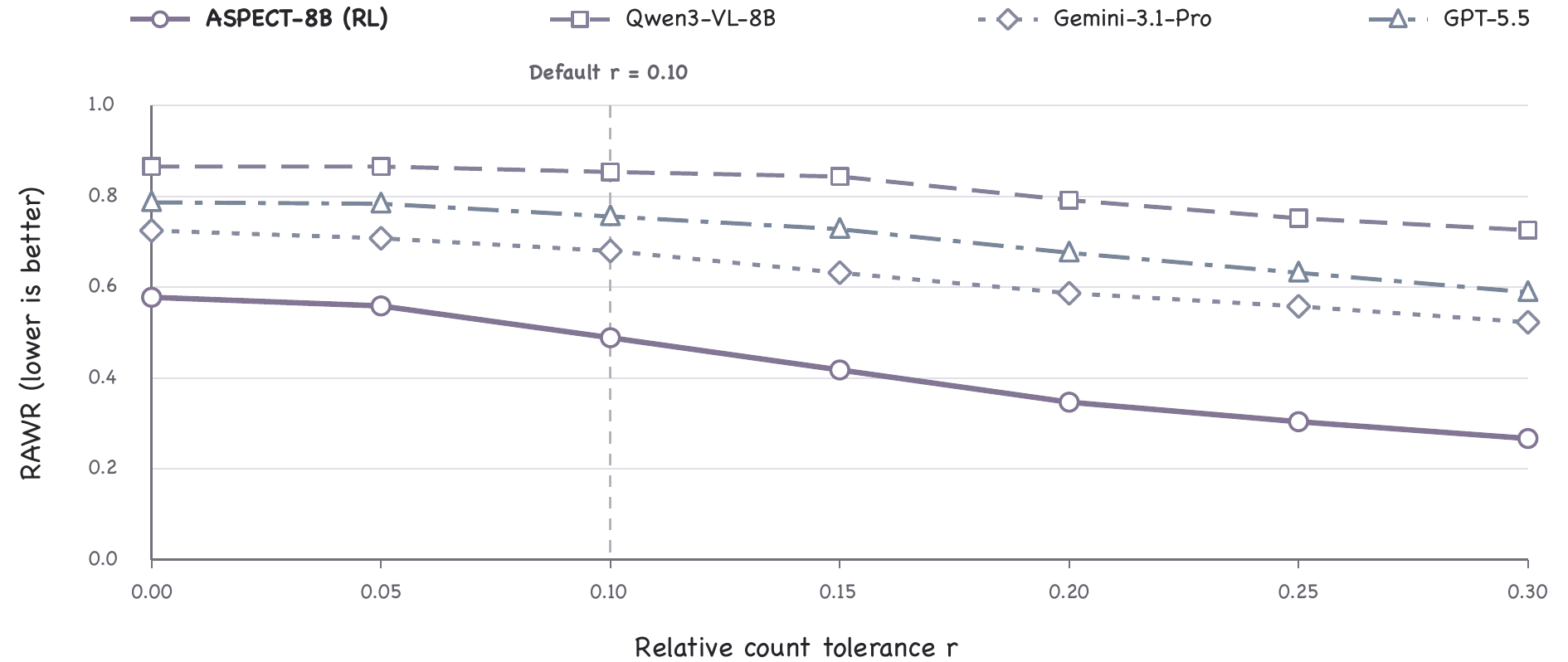}
\caption{RAWR under varying relative count tolerance $r$, with an absolute tolerance floor of one nucleus. The dashed line marks the main evaluation setting, $r=0.10$. Each model's eligible sample set remains fixed throughout the scan. Lower is better.}
\label{fig:count_tolerance}
\end{figure*}

Figure~\ref{fig:count_tolerance} shows that ASPECT achieves the lowest RAWR at all seven tested tolerances. Even under the strictest setting, its RAWR is 0.578, compared with 0.725 for Gemini-3.1-Pro. Their gap increases from 0.147 at $r=0$ to 0.256 at $r=0.30$. The consistent ranking across the scan shows that ASPECT's advantage in reported count accuracy extends beyond the default tolerance.

\section{Complete Case Studies and Failure Analysis}
\label{app:case_studies}

\subsection{Complete Responses for Main-Text Examples}
\label{app:main_text_cases}

\begin{figure*}[h]
\centering
\includegraphics[width=\textwidth]{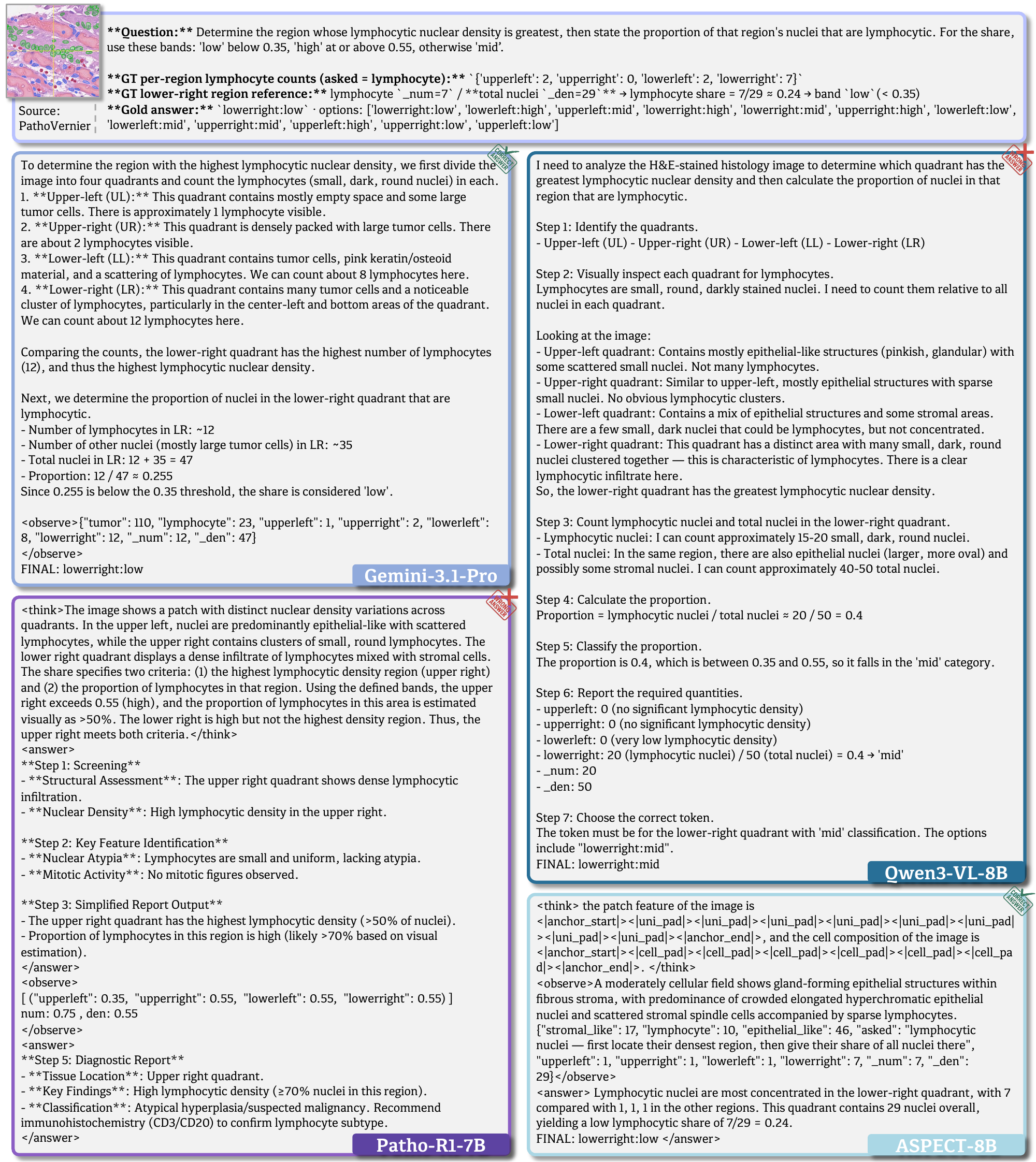}
\caption{Complete responses for the PathoVernier multi-step example. Gemini reaches the correct answer with inaccurate counts, whereas ASPECT recovers the reference counts of 7 lymphocytes among 29 nuclei in the selected region.}
\label{fig:full_case_pathovernier}
\end{figure*}

\begin{figure*}[p]
\centering
\includegraphics[width=\textwidth]{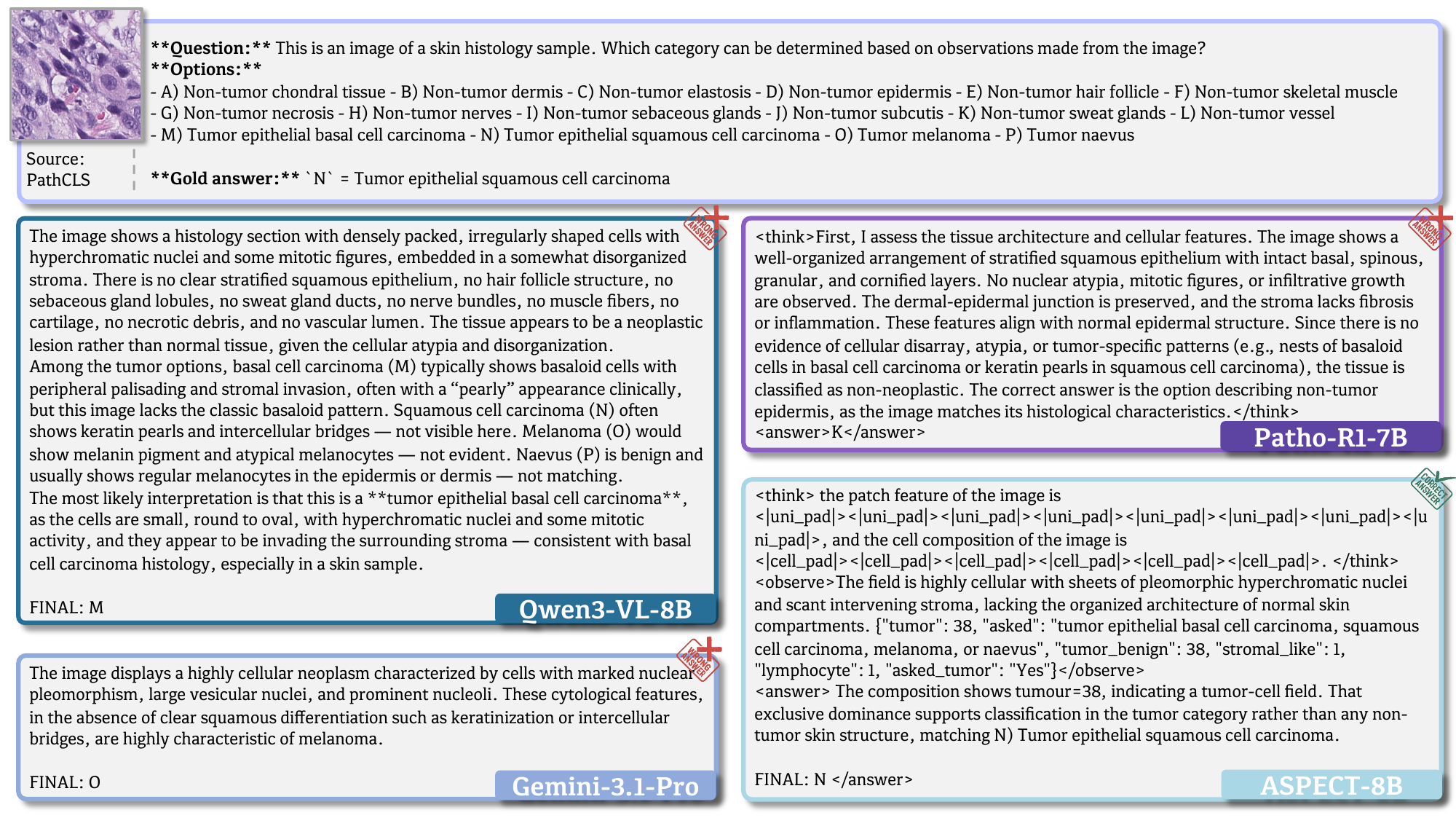}
\caption{Complete responses for the PathCLS example. ASPECT selects the reference class, squamous cell carcinoma (N), while the three comparison models select other classes.}
\label{fig:full_case_pathcls}
\end{figure*}

Figures~\ref{fig:full_case_pathovernier} and~\ref{fig:full_case_pathcls} provide the complete responses for the two examples discussed in the main-text qualitative comparison. Each figure includes the question, reference answer, and outputs from ASPECT, Qwen3-VL-8B, Gemini-3.1-Pro, and Patho-R1-7B.

\paragraph{PathoVernier: Correct answers and their quantitative basis.}
The reference answer selects the lower-right region and assigns a low lymphocyte proportion, based on 7 lymphocytes among 29 nuclei. Gemini reaches the correct answer using inaccurate counts of 12 and 47: their ratio falls in the same band despite the counting errors. Qwen3-VL-8B also selects the correct region, but its estimate of 20 out of 50 nuclei leads to the incorrect mid band. ASPECT recovers the reference counts for the selected region and derives the correct low band from $7/29\approx0.24$. This example illustrates why final-answer accuracy alone cannot distinguish accurate measurements from counting errors that preserve the answer category.

\paragraph{PathCLS: Tissue and lesion classification.}
For the skin histology example, ASPECT selects the reference class, squamous cell carcinoma (N), while Qwen3-VL-8B selects basal cell carcinoma (M) and Gemini selects melanoma (O). Patho-R1 describes non-tumor epidermis in its reasoning but outputs K, which denotes non-tumor sweat glands. The complete responses expose differences in both class selection and the correspondence between the written rationale and the selected option.

\subsection{Additional PathoVernier Examples}
\label{app:additional_pathovernier}

Figures~\ref{fig:case_region_compare}--\ref{fig:case_density_selection} supplement the multi-step example in Appendix~\ref{app:main_text_cases} with region comparison, cell-type comparison, and region selection. The complete responses show how count estimates and their subsequent interpretation affect the final answer.

\paragraph{Region comparison.}
In Figure~\ref{fig:case_region_compare}, the reference tumor counts are 15 in the top half and 2 in the bottom half, giving the answer \texttt{much\_more}. ASPECT reports 16 and 3 and reaches the same comparison category. Gemini reports 12 and 8, leading to the incorrect \texttt{comparable} answer. Qwen3-VL-8B exhibits a further discrepancy: its reasoning uses $6.5/3.5$, but its final reported counts are 6 and 3. The latter imply \texttt{much\_more} at the stated threshold, although its final answer remains \texttt{comparable}.

\paragraph{Cell-type comparison.}
Figure~\ref{fig:case_type_compare} asks whether stromal and inflammatory nuclei have comparable abundance. The reference counts are 57 and 44. Gemini estimates 2 and 65 and answers \texttt{much\_fewer}, whereas Qwen3-VL-8B reports 17 and 6 and answers \texttt{much\_more}. ASPECT reports 54 and 50 and correctly selects \texttt{comparable}. The opposing baseline conclusions arise from markedly different estimates of the same two cell populations.

\paragraph{Region selection.}
In Figure~\ref{fig:case_density_selection}, the reference stromal counts across four equal-area horizontal bands are $(16,4,1,1)$, making the topmost band, \texttt{r1}, the correct answer. ASPECT reports $(16,5,2,1)$ and selects \texttt{r1}, recovering the dominant concentration in the top band. Patho-R1's reasoning also points to \texttt{r1}, although its reported counts of $(2,0,0,1)$ substantially underestimate the stromal population. Gemini and Qwen3-VL-8B instead select \texttt{r2} and \texttt{r4}, respectively. This example further separates correct regional selection from accurate measurement.

\subsection{Failure Analysis}
\label{app:failure_analysis}

Figure~\ref{fig:failure_cases} presents two ASPECT failures in which the final answers follow the reported counts, but those counts misrepresent the regional composition.

\paragraph{Incorrect regional distribution despite a near-correct total.}
In the CoNSeP example (right), ASPECT reports 47 tumor nuclei across the image, close to the reference total of 46. Its regional counts, however, are $(0,4,7,36)$ rather than $(0,8,28,10)$. Selecting the largest reported count consequently yields \texttt{r4} instead of the reference region \texttt{r3}. The response is internally consistent, but the near-correct total does not ensure accurate regional measurement. Across 43 failed strip-based density-selection questions, 27 (62.8\%) select a band adjacent to the reference band.

\paragraph{Denominator overestimation after correct region selection.}
In the Lizard example (left), ASPECT correctly identifies the upper-right quadrant and estimates 19 stromal nuclei, close to the reference count of 18. It nevertheless reports 53 total nuclei in that region instead of 31, reducing the estimated stromal proportion from the reference $18/31\approx0.58$ to $19/53\approx0.36$. This changes the answer from \texttt{mid} to \texttt{low} at the 0.50 threshold. The case passes the region-only CA check but fails Acc, which evaluates the complete region-and-band answer. Among 44 multi-step questions with correct region selection but an incorrect proportion band, the median predicted-to-reference denominator ratio is 1.72, close to this example's $53/31\approx1.71$.

\section{Limitations and Future Work}
\label{app:limitations}

\paragraph{Spatial and clinical scope.}
This study focuses on quantitative visual reasoning within H\&E image patches. Pathological interpretation also draws on tissue architecture over larger fields, relationships between spatially separated regions, and patient-level context. Extending the present framework to whole-slide analysis would require coordinating local measurements with region selection and broader tissue organization. An important direction is to study how verifiable observations at different spatial scales can support slide-level assessments and clinically relevant endpoints.

\paragraph{Task and annotation scope.}
PathoVernier evaluates cell composition through a shared taxonomy and questions with explicit quantitative reference answers. This formulation makes intermediate measurements assessable across datasets, while its scope remains tied to the categories and quantities supported by the source annotations. Morphological attributes, interactions between cell populations, and richer descriptions of tissue organization offer complementary dimensions of pathology understanding. Future work could develop reference annotations and evaluation criteria for these dimensions, extending the connection between visual evidence and model conclusions beyond the count-based tasks studied here.

\begin{figure*}[p]
\centering
\includegraphics[width=\textwidth]{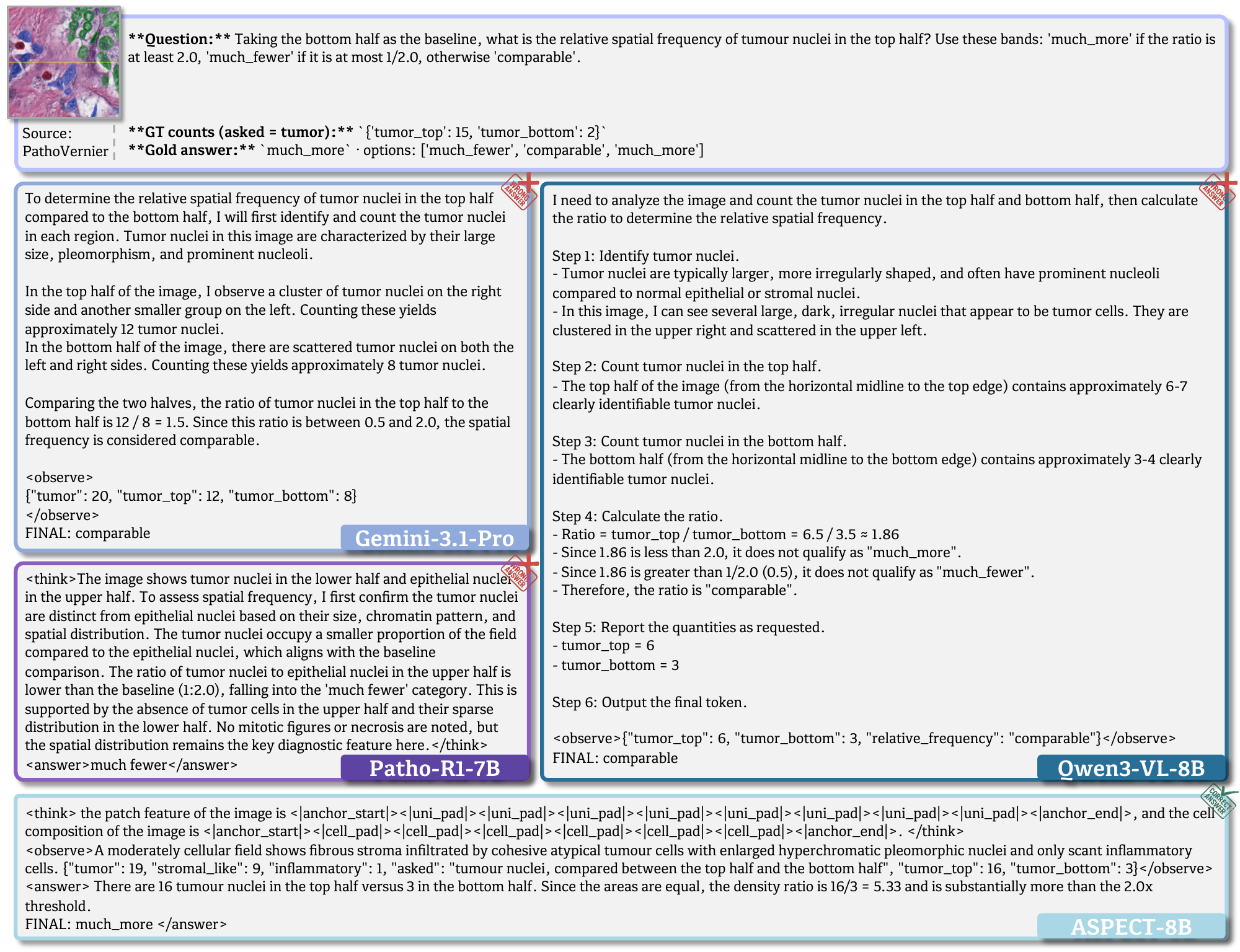}
\caption{Regional comparison of tumor nuclei. ASPECT selects the correct ratio band; the baseline responses illustrate errors in count estimation and in translating reported counts into the final answer.}
\label{fig:case_region_compare}
\end{figure*}

\begin{figure*}[p]
\centering
\includegraphics[width=\textwidth]{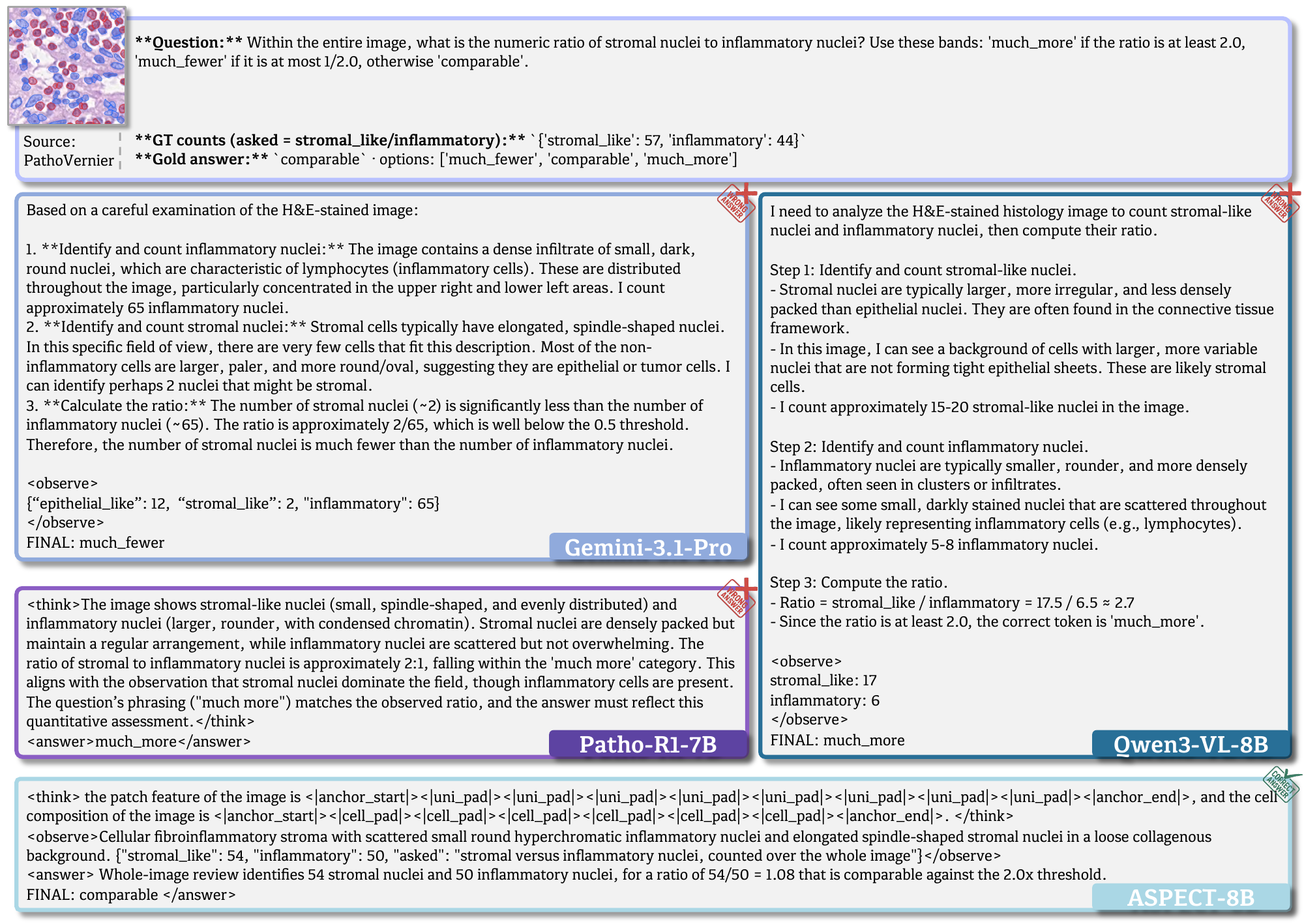}
\caption{Whole-image comparison of stromal and inflammatory nuclei. ASPECT correctly selects \texttt{comparable}, while Gemini and Qwen3-VL-8B reach opposite incorrect conclusions from their count estimates.}
\label{fig:case_type_compare}
\end{figure*}

\begin{figure*}[p]
\centering
\includegraphics[width=\textwidth]{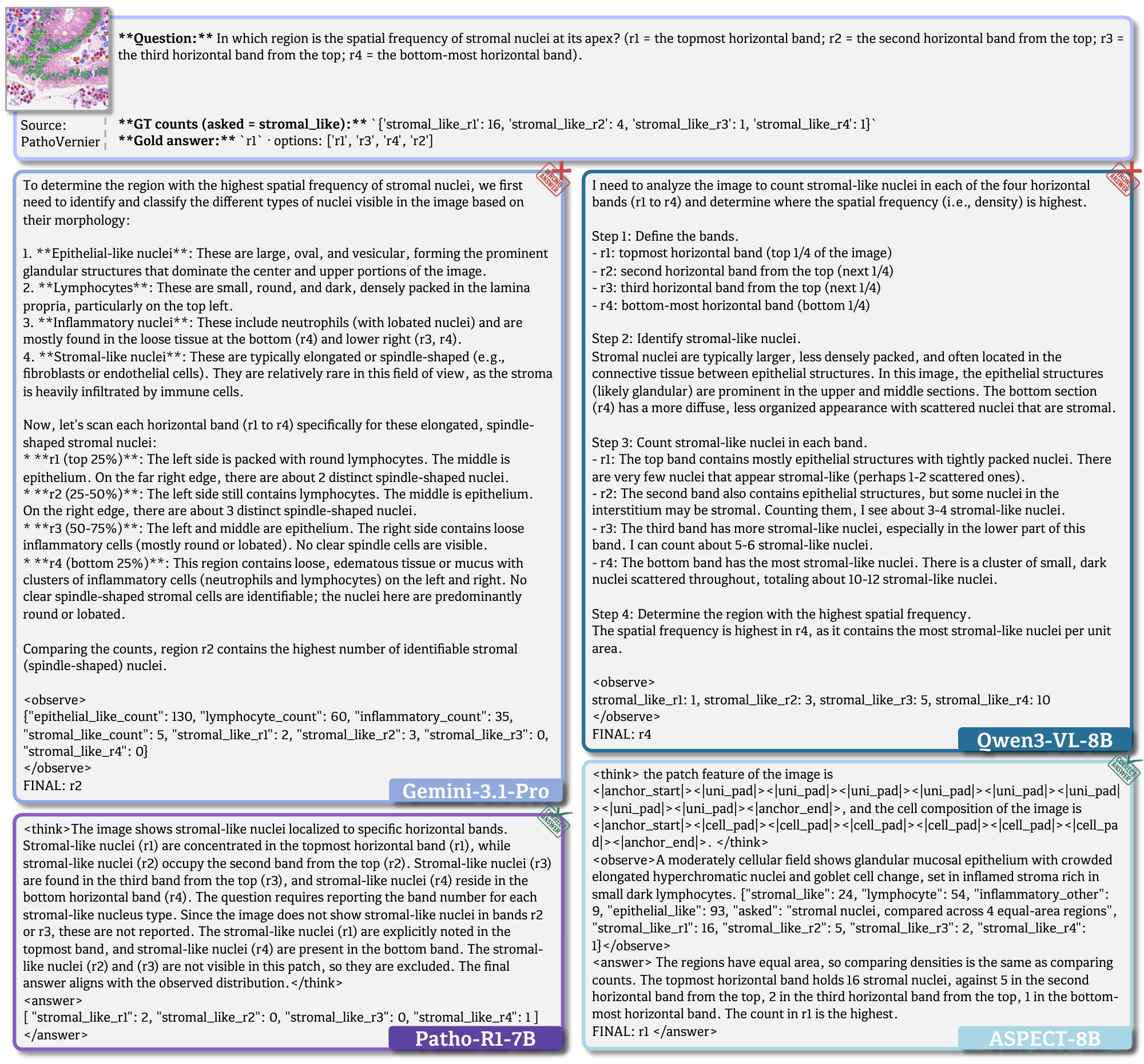}
\caption{Stromal-density selection across four equal-area horizontal bands. ASPECT identifies the reference region, \texttt{r1}, while Gemini and Qwen3-VL-8B select \texttt{r2} and \texttt{r4}, respectively.}
\label{fig:case_density_selection}
\end{figure*}

\begin{figure*}[p]
\centering
\includegraphics[width=\textwidth]{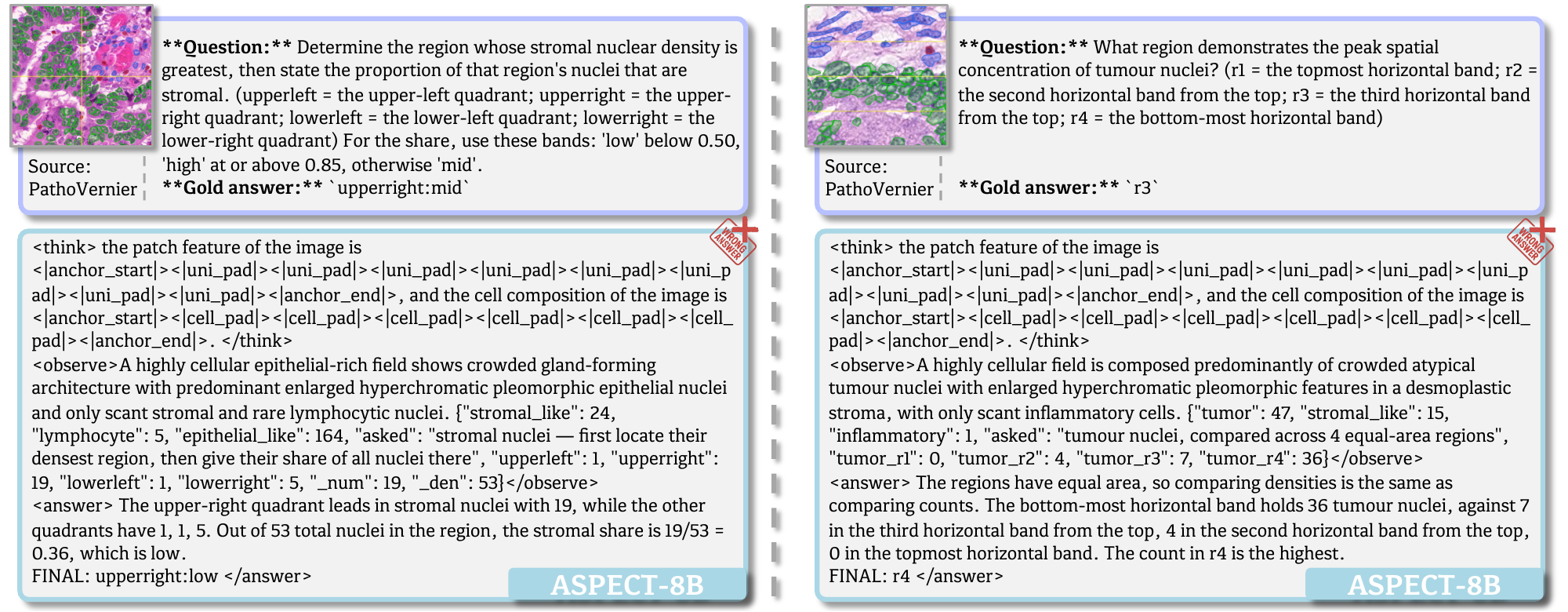}
\caption{ASPECT failure cases. Left: correct region selection followed by an incorrect proportion band due to denominator overestimation. Right: a near-correct whole-image count accompanies an incorrect regional distribution and region choice. Both final answers are consistent with the reported counts.}
\label{fig:failure_cases}
\end{figure*}

\end{document}